\documentclass{article} % For LaTeX2e
\usepackage{iclr2027_conference,times}
\iclrfinalcopy
\usepackage{hyperref}
\usepackage{url}
\usepackage{amsmath,amssymb}
\usepackage{booktabs}
\usepackage{graphicx}
\usepackage{multirow}
\usepackage{wrapfig}
\usepackage[table]{xcolor}
\usepackage{caption}
\usepackage[most]{tcolorbox}
\usepackage{needspace}

\newcommand{\samplebox}[3]{%
  \Needspace{18\baselineskip}
  \begin{tcolorbox}[
    enhanced,
    skin=enhanced jigsaw,
    breakable,
    width=0.97\linewidth,
    colback=white,
    colframe=black,
    boxrule=0.5pt,
    arc=0pt,
    outer arc=0pt,
    left=5pt,
    right=5pt,
    top=5pt,
    bottom=5pt,
    boxsep=0pt,
    title={\textbf{#1}},
    coltitle=black,
    colbacktitle=black!8,
    colframe=black,
    fontupper=\ttfamily\small\raggedright,
    fonttitle=\normalfont,
    toptitle=2pt,
    bottomtitle=2pt,
    lefttitle=5pt,
    righttitle=5pt,
    titlerule=0.4pt,
  ]
  \if\relax\detokenize{#2}\relax\else #2\par\vspace{10pt}\fi
  #3
  \end{tcolorbox}}

\title{One-Step Text Generation by Seq-Drifting}

\author{
\hspace{-0.3em}Jiayu Liu\textsuperscript{1}\hspace{0.3em}
Ruizhi Zhao\textsuperscript{3}\hspace{0.3em}
Tianyun Ji\textsuperscript{4}\hspace{0.3em}
Yee Whye Teh\textsuperscript{5}\hspace{0.3em} 
Zhenya Huang\textsuperscript{4}\hspace{0.3em}
Ning Miao\textsuperscript{1,2*}
}

\begin{document}

\vspace*{-0.2in}
\maketitle

% Author details without automatic title-footnote symbols.
\begingroup
\renewcommand{\thefootnote}{}
\footnotetext[0]{

\textsuperscript{1}Hong Kong Institute of AI for Science, City University of Hong Kong,~
\textsuperscript{2}Department of Data Science, City University of Hong Kong,~
\textsuperscript{3}Shenzhen University,~
\textsuperscript{4}State Key Laboratory of Cognitive Intelligence, University of Science and Technology of China,~
\textsuperscript{5}Department of Statistics, University of Oxford.}
\footnotetext[0]{\textsuperscript{*}Correspondence to: Ning Miao (ningmiao@cityu.edu.hk)}
\endgroup

\begin{abstract}
Diffusion language models generate tokens in parallel, but still require multiple sampling steps to produce high-quality text. In this paper, we investigate whether a model can learn to generate an entire sequence in a single forward pass. We propose \textit{Seq-Drifting}, a framework that applies drifting to sequences of continuous token representations and directly learns a map from Gaussian noise to text. Unlike existing DLMs that derive the supervision at each position directly from the observed token in a training sequence, Seq-Drifting builds a drift field whose position-wise targets adapt to the current generated context. At each position, attraction guides each output toward a set of contextually plausible token representations, while repulsion discourages collapse across samples and repetition within sequences. The generator learns from this drift field during training and produces all tokens jointly in one pass at inference. Empirically, Seq-Drifting achieves a generative perplexity of $36.01$ on LM1B in one step and performs competitively in both unconditional and conditional generation. Experiments on translation, summarization, and reasoning further demonstrate the applicability of the framework beyond open-ended generation. These results suggest that drifting over token sequences offers a promising path toward effective one-step text generation. Our code is available at \url{https://github.com/jyliu-ai/Seq-Drifting}.
\end{abstract}

\begin{figure}[h]
\centering
\includegraphics[width=\linewidth]{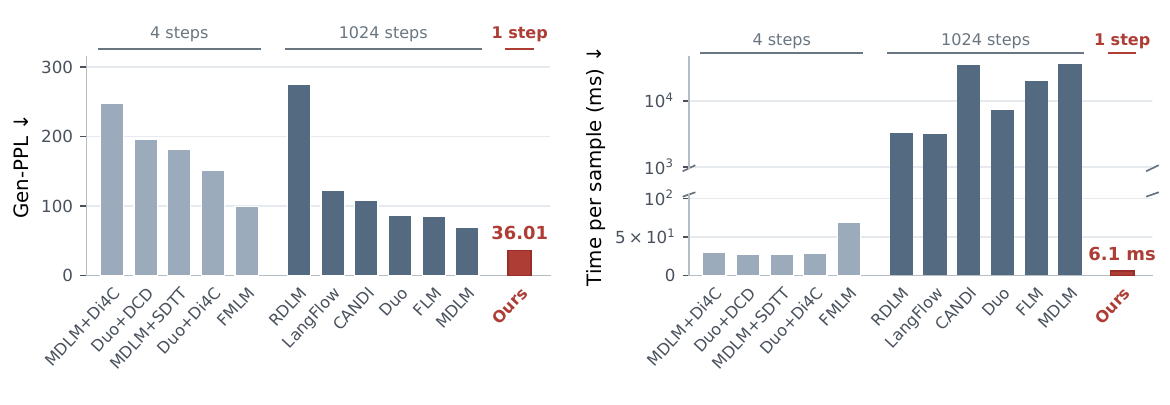}
\setlength{\abovecaptionskip}{-15pt}
\caption{\textbf{Seq-Drifting} achieves lower generative perplexity than prior
few-step distilled DLMs, and is on par with or better than multi-step DLMs, with a $4$--$11\times$ speedup over the few-step baselines and a $900$--$2500\times$ speedup over the
multi-step ones in wall-clock time. (Gen-PPL under GPT-2 Large; dataset: LM1B, $T{=}128$.
Detailed comparison in Figure~\ref{fig:main} and Table~\ref{tab:main_multi}.)}
\label{fig:teaser}
\vspace{-5pt}
\end{figure}

% =====================================================================
\section{Introduction}
Autoregressive (AR) language models are the dominant paradigm for text generation and have shown strong capabilities across a wide range of domains~\citep{radford2019language,brown2020language}. At inference, however, they generate one token at a time, making sequential decoding a fundamental latency bottleneck that grows with sequence length. Non-autoregressive alternatives such as diffusion language models (DLM) generate multiple tokens in parallel~\citep{sahoo2024simple,li2022diffusion,hu2026elf,lee2026one}, but they still rely on multiple refinement steps to produce high-quality samples. Distillation methods such as \mbox{SDTT~\citep{deschenaux2025beyond}}, Di4C~\citep{hayakawa2025distillation}, and FMLM~\citep{lee2026one} reduce the number of steps, but generation quality remains limited at small sampling budgets. A key question, then, is whether we can directly learn high-quality one-step generation through a single forward pass from noise to text.

Most generative models can be seen as a map from a noise vector to data. Because this map is highly multimodal, previous methods often need to decompose it into many small steps. Drifting~\citep{deng2026generativemodelingdrifting} offers a different route, learning the map directly as a single forward pass in continuous domains. Rather than learning a trajectory that is unrolled at inference, a drifting model evolves its pushforward map during training. As the parameters are updated, each noise sample is carried a little closer toward the data distribution, a movement governed by a drifting field that vanishes once the generated and data distributions match. At convergence the map can reach a fixed
point, so a single forward pass can turn noise into a sample without an iterative sampler at inference.

Although drifting has shown promising results on images, transferring it to text is far from straightforward. 
%\TODO{Ning: The sample in vanilla drifting is image, the sample in text-drifting is a sequence, instead of tokens.}
%The image drift presumes a continuous space in which every point is a valid sample and a small displacement yields a small, coherent change, so moving a sample toward nearby data is usually well defined. 
A text sample is a sequence in which every position contains a token drawn from a finite vocabulary.
%, and it meets neither condition. 
%Text meets neither condition. 
Since valid text sequences are composed of discrete token choices, a continuous drift step will generally land between valid token embeddings rather than on one. The drift may therefore operate in regions containing no valid token.
The space is also semantically non-smooth, as a single substituted token can overturn the meaning of a sentence or make it incoherent, which means nearness in embedding space does not necessarily imply nearness in meaning.
% and the direction a sample should take is ill-posed. 
%The supervision compounds the problem, since a position admits many valid tokens yet the data expose only one, leaving the field without the local set of targets it needs to define a direction.
One might hope to sidestep these issues by drifting a single sentence-level embedding, where the space
is continuous. However, this only relocates the difficulty, because mapping such a vector back to a faithful
token sequence is itself a hard sequence-decoding problem, and a single vector exposes none of the
per-position targets a drift step needs.

In this work we introduce \textbf{Seq-Drifting}, which adapts drifting to sequences of continuous token representations and learns the noise-to-text map in a single forward pass. Our core idea is to provide each position with supervision from a context-dependent set of plausible tokens. This support set anchors the drift to actual vocabulary tokens and identifies which choices are compatible with the current generated context. Seq-Drifting then attracts each output toward these tokens, allowing different noise inputs to select different valid alternatives without pulling them all toward an average representation. Because the support tokens depend on the generated context, the supervision couples token choices across positions and guides the generator toward coherent sequences. 
%The sets can be obtained through corpus retrieval or scoring by a frozen language model. 
%Repulsion complements this supervision by discouraging collapse across samples and repetition within a sequence. To keep these forces aligned, Seq-Drifting places all continuous representations on a unit sphere, where attraction and repulsion change token directions without unbounded growth in embedding norms.

Seq-Drifting establishes a new framework for one-step text generation and brings three advantages. First, it generates a sample in a single forward pass, so its inference cost does not grow with a step budget as in multi-step diffusion and few-step distillation, while matching their quality at a fraction of the inference cost.
Second, it addresses the ambiguity of one-step generation without relying on iterative refinement. Rather than assigning each noise input a unique target sequence, it allows different noise to naturally settle on their own coherent continuations while preserving multiple generation modes. Finally, it unifies two ways of learning the one-step map by constructing support sets either from corpus retrieval or scoring by a frozen language model, which opens the possibility of turning an existing autoregressive language model into a one-step generator.

Empirically, we show that Seq-Drifting combines competitive generation quality with substantially lower inference cost. For example, in Figure~\ref{fig:teaser}, it reaches a generative perplexity of $36.01$ with an $11.2\times$ wall-clock speedup over the strongest four-step distilled baseline, even outperforming several DLMs that use $1024$ sampling steps. This advantage extends to prefix-conditioned generation, such as machine translation, summarization, and reasoning, which broadens its applicability beyond language continuation. Overall, these results demonstrate the feasibility of directly learning a one-step text generator and point to a new direction for diffusion language models.

% =====================================================================
\section{Related Work}
Diffusion language models (DLMs) provide an iterative non-autoregressive alternative to the token-by-token decoding of autoregressive models \citep{radford2019language,brown2020language}, and can be broadly categorized into discrete and continuous formulations~\citep{li2025survey}. 

Discrete DLMs operate directly over categorical tokens. D3PM~\citep{austin2021structured} provides a general framework for discrete corruption, while MDLM~\citep{sahoo2024simple} uses masking to learn token recovery from partially observed sequences. LLaDA~\citep{nie2026large} further demonstrates masked diffusion at the scale of large language models. Unlike standard irreversible unmasking, the uniform-state formulation of Duo~\citep{sahoo2025diffusion} allows previously generated tokens to be revised. Block diffusion methods~\citep{arriola2025block,arriola2026encoder} instead combine parallel generation within blocks with autoregressive conditioning across blocks. However, parallel per-position predictions do not by themselves capture the joint uncertainty of tokens updated simultaneously, making inter-token dependencies particularly important when sampling is compressed into large jumps.

Continuous DLMs instead model language generation in a continuous representation space. Diffusion-LM~\citep{li2022diffusion} and DiffuSeq~\citep{gong2022diffuseq} operate on token embeddings, whereas SSD-LM~\citep{han2023ssd} and TESS~\citep{mahabadi2024tess} use simplex-based representations. Beyond the state space, an important distinction is how discrete supervision constrains the trajectory. FLM~\citep{lee2026one} learns continuous denoising over one-hot encodings with token-level cross-entropy, and LangFlow~\citep{chen2026langflow} connects embedding-space diffusion to Flow Matching through Bregman divergence. ELF~\citep{hu2026elf} instead reserves token-level supervision for the final decoding step, allowing the preceding dynamics to remain in continuous embedding space. These advances improve continuous generation, but an effective trajectory still requires tens or even hundreds of sampling steps at inference.

Few-step methods address this problem by learning to traverse or replace an existing sampling trajectory. SDTT~\citep{deschenaux2025beyond} compresses denoising through self-distillation, Di4C~\citep{hayakawa2025distillation} explicitly addresses the dimensional correlations lost in factorized reverse transitions, and FMLM~\citep{lee2026one} distills a continuous flow into a map between distant timepoints. Drifting~\citep{deng2026generativemodelingdrifting} provides a different training formulation, evolving the generator's outputs through attraction and repulsion while retaining a single forward pass at inference. Seq-Drifting extends this formulation to token sequences. Context-dependent, per-position supervision supplies plausible token targets for its current outputs. The support sets can draw on corpus retrieval or a frozen language model, while the objective directly trains the one-step map without reproducing a diffusion or flow trajectory.
% =====================================================================
\section{Preliminaries}
\label{sec:prelim}
We consider a one-step generator $G_\theta$ that maps a
noise input $\epsilon\sim p_\epsilon$ to an output $\boldsymbol{x}=G_\theta(\epsilon)$ with
distribution $q_\theta$, and training seeks $q_\theta\approx p_{\mathrm{data}}$. Since the
parameters are updated step by step, the same noise $\epsilon$ produces a slightly different output
after each step. The displacement
\begin{equation}
\Delta\boldsymbol{x}_i \;=\; G_{\theta_{i+1}}(\epsilon)-G_{\theta_i}(\epsilon)
\end{equation}
comes entirely from the parameter update, and is called the \emph{drift} of $\boldsymbol{x}$ at
iteration $i$. A \emph{drifting field} $V_{p,q}$~\citep{deng2026generativemodelingdrifting}
prescribes this drift sample by sample, with $p$ fixed to the target $p_{\mathrm{data}}$ and $q$ to
the current model $q_{\theta_i}$,
\begin{equation}
\boldsymbol{x}_{i+1}=\boldsymbol{x}_i+V_{p_{\mathrm{data}},\,q_{\theta_i}}(\boldsymbol{x}_i).
\end{equation}
The field vanishes when the two distributions agree, $V_{p,p}=\boldsymbol{0}$, so
$q_\theta=p_{\mathrm{data}}$ is a fixed point of training. Regressing each output onto its drifted target
with a stop gradient turns this into the loss,
\begin{equation}
\mathcal{L}(\theta)=\mathbb{E}_\epsilon\big\|\,G_\theta(\epsilon)-\operatorname{sg}\!\big(G_\theta(\epsilon)+V(G_\theta(\epsilon))\big)\big\|^2 ,
\label{eq:drift_loss}
\end{equation}
whose value equals $\mathbb{E}_\epsilon\|V(G_\theta(\epsilon))\|^2$, so minimizing it drives the
drift toward zero.

A concrete field comes from a kernel mean-shift. For a sample $\boldsymbol{x}$, a positive set
$\mathcal{P}$ drawn from $p_{\mathrm{data}}$ attracts and a negative set $\mathcal{N}$ drawn from
$q_\theta$ repels,
\begin{equation}
V(\boldsymbol{x})=\underbrace{\sum_{\boldsymbol{y}\in\mathcal{P}}\tilde{k}(\boldsymbol{x},\boldsymbol{y})\,(\boldsymbol{y}-\boldsymbol{x})}_{\text{attraction}}
\;-\;
\underbrace{\sum_{\boldsymbol{y}\in\mathcal{N}}\tilde{k}(\boldsymbol{x},\boldsymbol{y})\,(\boldsymbol{y}-\boldsymbol{x})}_{\text{repulsion}},
\label{eq:meanshift}
\end{equation}
where $\tilde{k}$ is a similarity kernel normalized over each set. The attraction moves
$\boldsymbol{x}$ toward the data it already resembles and the repulsion moves it off the other
samples, and the two cancel once the generated and data distributions agree.

\paragraph{Defining the drift target.}
The field in Eq.~(\ref{eq:meanshift}) specifies how a generated sample should move during training.
Positive samples provide the destinations for attraction, while other generated samples provide
the reference points for repulsion. The similarity kernel determines which samples contribute
most strongly to each force. Combining the forces gives a displacement, which defines the
stop-gradient target $\boldsymbol{x}+V(\boldsymbol{x})$ in Eq.~(\ref{eq:drift_loss}).
The generator learns to approach this target through parameter updates. Adapting drifting to a
new domain therefore requires suitable positive and negative samples, a geometry for comparing
them, and a rule for converting the field into a training target.

\paragraph{Difficulties in applying drifting to text.}
A text sequence is composed of discrete token choices,
which creates three difficulties for drifting. First, valid outputs lie in a discrete and
semantically non-smooth space. A continuous displacement can change the decoded token, but
proximity between embeddings does not guarantee that the new token fits the surrounding sequence.
Second, the tokens that
form a coherent continuation depend on the preceding sequence. However, a training example supplies only the continuation observed after its own context, which may differ substantially from the generator's current context. Using that observed continuation directly as a drift target can therefore yield a weak or misaligned update.
Third, under cosine decoding, radial separation can increase Euclidean distances without changing
token identity. Unconstrained repulsion can therefore spend its force on growing norms while
producing no additional token-level diversity.
%These considerations motivate context-dependent token support sets and a spherical representation.
%In \S\ref{sec:method}, we address these issues through context-dependent token support sets and a spherical representation.

\begin{figure*}
  \centering
  \includegraphics[width=\linewidth]{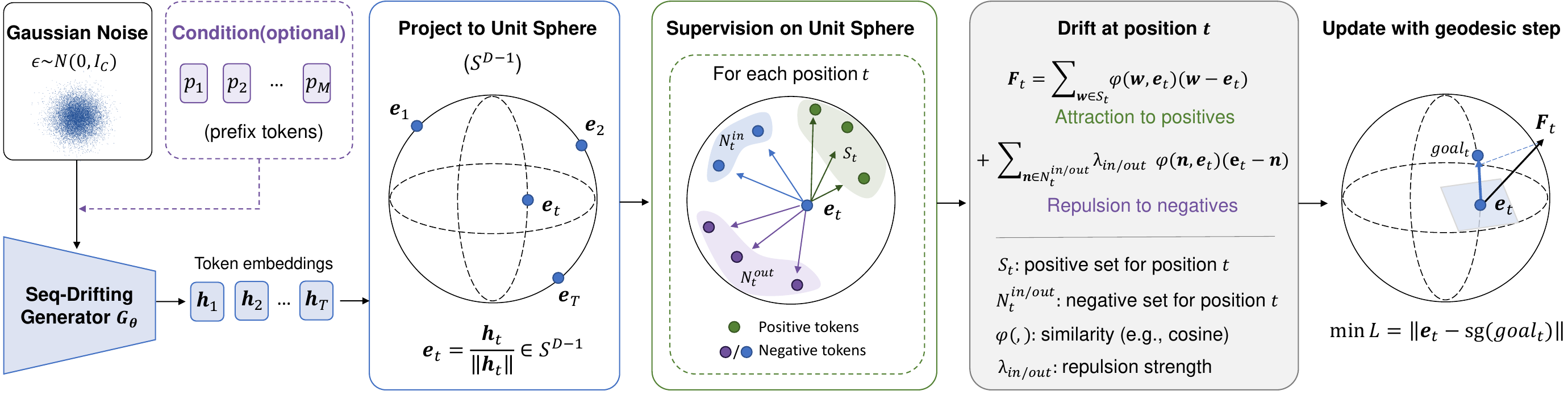}
  \caption{\textbf{Seq-Drifting.} In one forward pass the generator $G_\theta$ maps a noise vector
  $\epsilon\!\sim\!\mathcal{N}(0,I_C)$ (optionally with a prefix $p_1,\dots,p_M$) to token embeddings
  $\boldsymbol{h}_1,\dots,\boldsymbol{h}_T$, each projected onto the unit sphere as $\boldsymbol{e}_t=\boldsymbol{h}_t/\|\boldsymbol{h}_t\|\in S^{D-1}$.
  Supervision is applied \emph{per position}. A positive set $\mathcal{S}_t$ of plausible tokens
  attracts $\boldsymbol{e}_t$, while the negative sets $N_t^{\mathrm{out}}$ (other samples) and
  $N_t^{\mathrm{in}}$ (other positions of the same sample) repel it, weighted by a similarity kernel
  $\varphi$ and strengths $\lambda_{\mathrm{in}/\mathrm{out}}$. The two combine into the drift force
  $\boldsymbol{F}_t$; a single geodesic step along $\boldsymbol{F}_t$ gives the stop-gradient target $goal_t$,
  onto which the generator is regressed, $\mathcal{L}=\mathbb{E}\|\boldsymbol{e}_t-\mathrm{sg}(goal_t)\|^2$.}
  \label{fig:framework}
  \vspace{-10pt}
\end{figure*}

% =====================================================================
\section{Seq-Drifting}
\label{sec:method}
To address the difficulties above, Seq-Drifting represents a sequence with per-position token
representations and provides each position with a context-dependent set of plausible tokens.
These representations and targets are placed in a shared spherical space, where the resulting
drift remains anchored to valid vocabulary while the noise-to-text map is learned
directly. Figure~\ref{fig:framework}
summarizes the framework. We first describe the one-step generator and its representation space,
then introduce support-set construction, the attractive and repulsive forces, and the
training objective.

\subsection{Spherical one-step generator}
\label{sec:tdd_gen}
We generate a sequence of continuous representations, each one corresponding to a token. This allows
the generator to produce all positions jointly while exposing each token choice to local drift
supervision. Specifically, our generator $G_\theta$ maps a shared noise vector $\epsilon\in\mathbb{R}^{C}$
to $T$ token representations $\{\boldsymbol{h}_t\}_{t=1}^{T}$, with
$\boldsymbol{h}_t\in\mathbb{R}^{D}$, in one forward pass. Architectural details are given
in Appendix~\ref{sec:app_generator}.
%A bi-directional Transformer lets the positions interact before their tokens are decoded. 
Then, we recover tokens by cosine similarity to a frozen vocabulary embedding table. Since this
decoding rule depends only on direction, we normalize both the output vectors $\boldsymbol{h}_t$ and the vocabulary embeddings $\boldsymbol{w}_v$, i.e.,
$\boldsymbol{e}_t=\boldsymbol{h}_t/\|\boldsymbol{h}_t\|\in S^{D-1}$ and
$\|\boldsymbol{w}_v\|=1$. Attraction, repulsion, and token recovery then depend on the same
directional geometry. The decoded token is
$\mathrm{tok}_t=\arg\max_v\boldsymbol{e}_t^{\top}\boldsymbol{w}_v$,
with non-text tokens excluded. This operation produces the final text at inference, requiring no separate vocabulary softmax.
% and the provisional tokens used to construct language-model support sets during training.

\subsection{Context-dependent token support}
\label{sec:tdd_support}
The main challenge in selecting positives is to provide useful guidance for the sequence
the generator currently produces. A reference sentence may differ from that sequence at many
positions, so its tokens need not be suitable targets for the corresponding positions. We instead
construct positive targets at each position, asking which tokens are plausible given the generated
context.

For position $t$, we collect the positive targets into a support set $\mathcal{S}_t$
(i.e., $\mathcal{P}$ in \S~\ref{sec:prelim}), with each
candidate represented by its unit embedding. The set supplies concrete token targets without
prescribing a single continuation. It separates contextual plausibility from the choice made
by an individual sample. The support set determines which tokens are eligible for attraction,
while the local kernel introduced next determines how strongly each candidate attracts the
current output.

There are two natural ways to form $\mathcal{S}_t$. The \emph{off-policy} route reads it from the
corpus, matching each generation against real sequences and taking the tokens they place at position
$t$, so the candidates come from observed continuations. The \emph{on-policy} route queries a frozen language model (hereafter, the \emph{support model}) on the generator's
current prefixes to propose contextually plausible next-token candidates, where one forward pass over the generated sequence
returns a next-token distribution $p_t$ at every position at once. $\mathcal{S}_t$ collects its
most probable tokens (the top $k$), so the target adapts to wherever the
current generator sits. The same construction also applies to a closed-source support model, which can
be queried multiple times to sample continuations from the generated context and form $\mathcal{S}_t$.
We empirically find the on-policy support the stronger of the two
(\S\ref{sec:exp_ablation}) and adopt it by default. The support model is queried only during training to construct the support set $\mathcal{S}_t$, and is not required at inference. The resulting supervision is set-valued and context-dependent rather than a fixed per-position target.
%, but the choice is orthogonal to the rest of the drift. 
%with probability above $0.01$
%Either set plugs into the same attraction of \S\ref{sec:tdd_attract} without changing how the force is computed. 
%The same mechanism serves unconditional generation and conditional generation with an observed context.

\subsection{Angular mode-seeking attraction}
\label{sec:tdd_attract}
The attraction pulls each position toward the supported token nearest its own embedding. We measure
closeness by angular distance $d(\boldsymbol{e}_t,\boldsymbol{w})=1-\boldsymbol{e}_t^{\top}\boldsymbol{w}$ and form a
low-temperature similarity kernel $\varphi$, normalized over the support set,
\begin{equation}
\varphi(\boldsymbol{w},\boldsymbol{e}_t)\;\propto\;\exp\!\Big(\!-\frac{d(\boldsymbol{e}_t,\boldsymbol{w})}{s_t\,\tau}\Big),
\qquad
\boldsymbol{u}_t=\!\!\sum_{\boldsymbol{w}\in\mathcal{S}_t}\!\varphi(\boldsymbol{w},\boldsymbol{e}_t)\,(\boldsymbol{w}-\boldsymbol{e}_t),
\qquad
s_t=\operatorname{std}_{\boldsymbol{w}\in\mathcal{S}_t} d(\boldsymbol{e}_t,\boldsymbol{w}),
\end{equation}
where $s_t$ is the standard deviation of $d(\boldsymbol{e}_t,\boldsymbol{w})$ for candidates at position $t$ and $\tau$ is a
temperature. Normalizing by $s_t$ accounts for each position's candidate spread, while $\tau<1$ sharpens the kernel around nearby candidates. It is therefore mode-seeking and concentrates the
weight on the near supported token, so different noise inputs commit to different tokens
rather than to the blurred centroid of $\mathcal{S}_t$. The mean-shift then gives the attractive
force $\boldsymbol{u}_t=\sum_{\boldsymbol{w}\in\mathcal{S}_t}\varphi(\boldsymbol{w},\boldsymbol{e}_t)(\boldsymbol{w}-\boldsymbol{e}_t)$.

\subsection{Per-position angular repulsion}
\label{sec:tdd_repel}
The repulsion stops distinct noise inputs from collapsing to the same output. It acts per position
against two negative sets, i.e., $\mathcal{N}\triangleq\mathcal{N}_t=\{N_t^{\mathrm{out}},N_t^{\mathrm{in}}\}$. The cross-sample set $N_t^{\mathrm{out}}$ contains the position-$t$ representations from the other samples in the batch. The intra-sequence set $N_t^{\mathrm{in}}$ contains the representations at the other positions of the same sequence, and
pushing away from it suppresses within-sentence repetition. Each uses the same angular kernel
$\varphi$ as the attraction,
\begin{equation}
\boldsymbol{r}^{\mathrm{out}}_t=\!\!\sum_{\boldsymbol{n}\in N_t^{\mathrm{out}}}\!\!\varphi(\boldsymbol{n},\boldsymbol{e}_t)\,(\boldsymbol{e}_t-\boldsymbol{n}),
\qquad
\boldsymbol{r}^{\mathrm{in}}_t=\!\!\sum_{\boldsymbol{n}\in N_t^{\mathrm{in}}}\!\!\varphi(\boldsymbol{n},\boldsymbol{e}_t)\,(\boldsymbol{e}_t-\boldsymbol{n}).
\end{equation}
%so each is a mean-shift away from its negatives, opposite in sign to the attraction. 
Both are then rescaled
to the magnitude of the attraction before the update weights them by $\lambda_{\mathrm{out}}$ and
$\lambda_{\mathrm{in}}$. Restricting the cross-sample set $N_t^{\mathrm{out}}$ to the samples that
share a context is studied in \S\ref{sec:exp_ablation}.

On the sphere, repulsion separates outputs by changing their directions rather than
increasing their norms. This prevents radial growth and aligns repulsion with cosine decoding.
Together with the bounded geodesic update in \S\ref{sec:tdd_update}, it defines the displacement
of each training target. 
%Repulsion strengths still balance diversity against attraction to plausible tokens, so spherical geometry alone does not guarantee stable training.

\subsection{Geodesic update and training}
\label{sec:tdd_update}
The update sums the attraction and the two repulsions into one drift force,
$\boldsymbol{F}_t=\boldsymbol{u}_t+\lambda_{\mathrm{out}}\,\boldsymbol{r}^{\mathrm{out}}_t+\lambda_{\mathrm{in}}\,\boldsymbol{r}^{\mathrm{in}}_t$,
projects it onto the tangent plane at $\boldsymbol{e}_t$ to drop the radial part,
$\boldsymbol{F}^{\prime}_t=\boldsymbol{F}_t-(\boldsymbol{F}_t^{\top}\boldsymbol{e}_t)\,\boldsymbol{e}_t$, and applies
it as one step along the great circle by the exponential map,
\begin{equation}
goal_t=\cos(\alpha)\,\boldsymbol{e}_t+\sin(\alpha)\,\frac{\boldsymbol{F}^{\prime}_t}{\|\boldsymbol{F}^{\prime}_t\|},
\qquad
\alpha=\min\!\big(\|\boldsymbol{F}^{\prime}_t\|,\ \alpha_{\max}\big).
\label{eq:geodesic}
\end{equation}
Here we use the norm of $\boldsymbol{F}^{\prime}_t$ as the arc length of the geodesic step, capped by
$\alpha_{\max}$ to prevent overshooting. Treating $goal_t$ as a fixed target during differentiation,
the loss of Eq.~(\ref{eq:drift_loss}) becomes the squared
distance $\mathbb{E}\,\|\boldsymbol{e}_t-\operatorname{sg}(goal_t)\|^2$, which is bounded because
both points lie on the sphere. Appendix~\ref{sec:app_fixedpoint} analyzes the conditions under which the field vanishes as the temperature approaches zero. Appendix~\ref{sec:app_algorithm} provides the training pseudocode for the on-policy
support-set implementation.

\paragraph{Conditional generation.}
For conditional generation, we additionally provide the generator with the conditioning token embeddings, which the output positions access through cross-attention. The support model also receives the conditioning sequence followed by the generated continuation to provide token support sets.
To stabilize learning this conditional map from scratch, we first train with an unconditional
warm-up. During this phase, the context is filled with a constant padding token
(e.g., \texttt{<|endoftext|>} for GPT-2) under normal attention, which keeps the conditioning
path active while the generator learns fluent text. We then replace the constant context with
the real prefix, allowing the generator to learn context dependence from this fluent initialization.
% =====================================================================
\section{Experiments}
\label{sec:exp}

\subsection{Setup}
\paragraph{Baselines.} Our main comparison is against the \emph{few-step distilled models} that compress a multi-step sampler into a handful of steps, including FMLM~\citep{lee2026one}, Duo+DCD~\citep{sahoo2025diffusion},
MDLM+SDTT~\citep{deschenaux2025beyond}, and Duo/MDLM+Di4C~\citep{hayakawa2025distillation}. We report
them at $1$, $2$, and $4$ steps. We also
compare against \emph{multi-step models} that generate through multiple (e.g., 1024) denoising steps,
namely FLM~\citep{lee2026one},
Duo~\citep{sahoo2025diffusion}, MDLM~\citep{sahoo2024simple}, CANDI~\citep{pynadath2026candi},
RDLM~\citep{jo2026continuous}, LangFlow~\citep{chen2026langflow}, and ELF~\citep{hu2026elf}.

\paragraph{Data and protocol.} We follow the evaluation protocol of \citet{lee2026one} and report at
two output lengths. In the length-$128$ setup the models are trained on One Billion
Word (LM1B)~\citep{chelba2013one}, and in the
length-$1024$ setup on OpenWebText~\citep{Gokaslan2019OpenWeb} using the corresponding benchmark
preprocessing. We draw $1024$
samples per configuration. For each method we report the number of forward passes per
sample (\textbf{Step}), the generative perplexity of the samples under GPT-2 Large
(\textbf{Gen-PPL}$\downarrow$), and their unigram entropy (\textbf{Entropy}).

\paragraph{Implementation details.}
%As described in Appendix~\ref{sec:app_generator}, our generator has approximately $98$M parameters for length $128$ and $187$M for length $1024$. 
We consider two implementations of on-policy support-set construction. 
For \textbf{Seq-Drifting$^{S}$}, we first train a GPT-2 from \emph{scratch} on the same corpus as the
baselines for $10^{5}$ steps, then freeze it and use it as the support model. This setting uses no additional text beyond the
baselines. For \textbf{Seq-Drifting$^{P}$}, we directly use a \emph{public} pretrained GPT-2
checkpoint to construct the support sets in the same way. 
The support set uses $k=64$, the attraction temperature is $\tau=0.3$, and the cross-sample and
intra-sequence repulsions use $\lambda_{\mathrm{out}}=5$ and $\lambda_{\mathrm{in}}=100$. The geodesic
step is capped at $\alpha_{\max}=1.5708$ radians. We train with AdamW (learning rate $2\times10^{-4}$,
weight decay $0$, cosine schedule) on $4$ H200 GPUs. Unconditional models are trained for $2\times10^{5}$ steps.
Conditional models use a $2\times10^{5}$-step
unconditional warm-up, followed by $1\times10^{5}$ conditional steps, keeping an exponential
moving average of the weights with decay $0.999$. Therefore, the total training step of Seq-Drifting is
shorter than the 1.0--1.5M-step recipes usually used by the baselines. The effective batch is $2048$ at length $128$ and
$128$ at length $1024$, the longer sequence being reduced to fit memory.

% Main unconditional result: few-step distilled table (left) + running-time/Gen-PPL figure (right),
% under one shared caption. Multi-step models are in the appendix (Table~\ref{tab:main_multi}).
\begin{figure}[t]
\centering
\begin{minipage}[c]{0.44\textwidth}
\centering
\scriptsize
\setlength{\tabcolsep}{1.6pt}
\renewcommand{\arraystretch}{1.2}
\begin{tabular}{l|c|cc|cc}
\toprule
\multirow{2}{*}{Method} & \multirow{2}{*}{Step} & \multicolumn{2}{c|}{LM1B} & \multicolumn{2}{c}{OWT} \\
\cmidrule(lr){3-4}\cmidrule(lr){5-6}
 & & Gen-PPL$\downarrow$ & Ent. & Gen-PPL$\downarrow$ & Ent. \\
\midrule
\multirow{3}{*}{FMLM}
 & 1 & 119.34 & 4.16 & 169.32 & 5.17 \\
 & 2 & 110.19 & 4.21 & 135.68 & 5.30 \\
 & 4 & 98.76  & 4.21 & 111.49 & 5.36 \\
\cmidrule(lr){1-6}
\multirow{3}{*}{Duo+DCD}
 & 1 & 1006.75 & 4.33 & 6330.89 & 5.98 \\
 & 2 & 466.46  & 4.24 & 983.80  & 5.52 \\
 & 4 & 195.52  & 4.23 & 252.04  & 5.49 \\
\cmidrule(lr){1-6}
\multirow{3}{*}{Duo+Di4C}
 & 1 & 292.94 & 3.79 & 370.51 & 3.92 \\
 & 2 & 247.69 & 3.87 & 210.22 & 4.63 \\
 & 4 & 150.67 & 4.00 & 154.67 & 4.85 \\
\cmidrule(lr){1-6}
\multirow{3}{*}{MDLM+SDTT}
 & 1 & 1099.64 & 4.23 & 1264.02 & 5.25 \\
 & 2 & 448.78  & 4.22 & 860.40  & 5.34 \\
 & 4 & 181.08  & 4.23 & 319.73  & 5.36 \\
\cmidrule(lr){1-6}
\multirow{3}{*}{MDLM+Di4C}
 & 1 & 1217.10 & 4.38 & 1272.27 & 5.28 \\
 & 2 & 621.59  & 4.37 & 742.99  & 5.34 \\
 & 4 & 247.32  & 4.00 & 226.95  & 5.37 \\
\midrule
\rowcolor{gray!12}
\textbf{Seq-Drifting$^{S}$} & \textbf{1} & \textbf{93.81} & \textbf{4.21} & \textbf{183.98} & \textbf{4.95} \\
\rowcolor{gray!12}
\textbf{Seq-Drifting$^{P}$} & \textbf{1} & \textbf{36.01} & \textbf{4.42} & \textbf{128.33} & \textbf{5.08} \\
\bottomrule
\end{tabular}
\end{minipage}\hfill
\begin{minipage}[c]{0.55\textwidth}
\centering
\includegraphics[width=\linewidth]{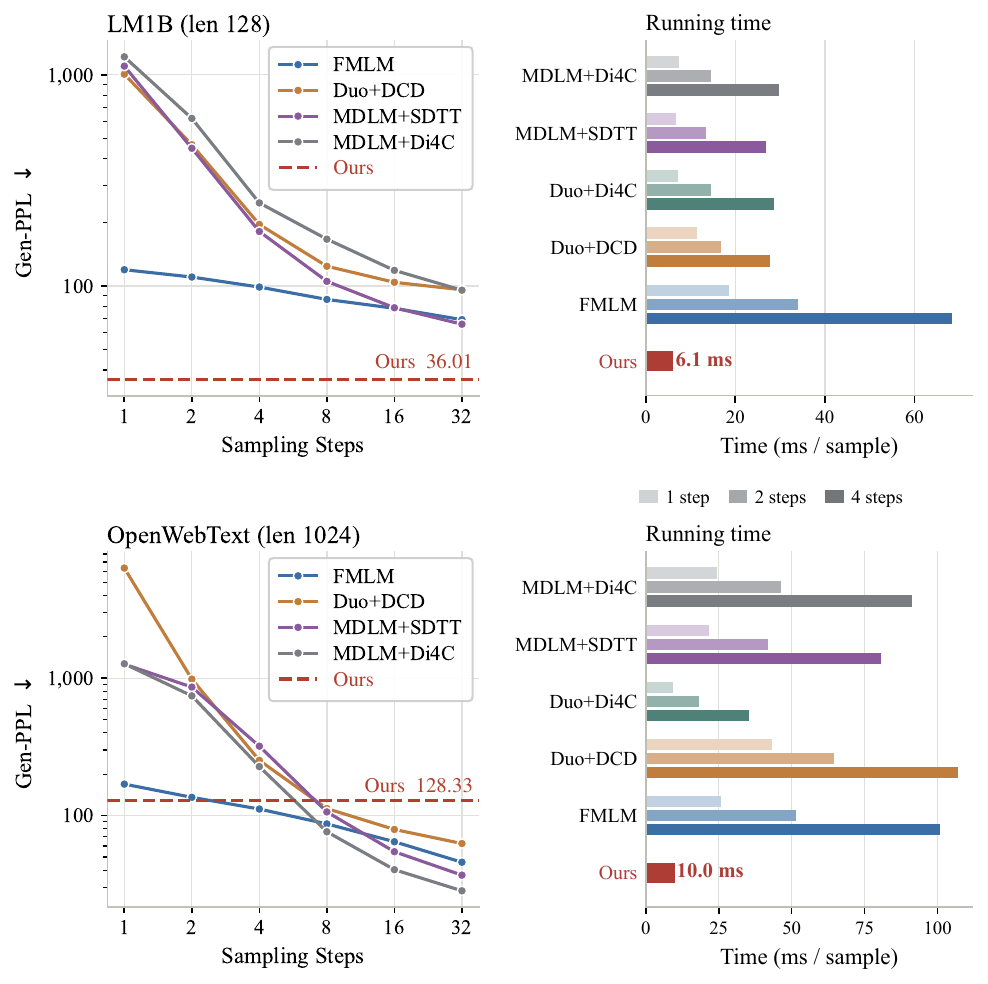}
\end{minipage}
\caption{\textbf{Unconditional generation.} \emph{Left:} Gen-PPL (under GPT-2 Large) and unigram
entropy for the few-step distilled models at $1$/$2$/$4$ steps and Seq-Drifting at a single pass, on LM1B
($T{=}128$) and OpenWebText ($T{=}1024$); multi-step baselines are in Table~\ref{tab:main_multi}.
\emph{Right:} Gen-PPL vs.\ sampling steps and wall-clock running time per sample, showing Seq-Drifting
achieves lower Gen-PPL in a single forward pass at a fraction of the cost.}
\label{fig:main}
\par\vspace{-10pt}
\end{figure}

\subsection{Unconditional generation}
\label{sec:exp_uncond}
Figure~\ref{fig:main} compares the few-step distilled models with Seq-Drifting at a single pass. The full
multi-step results are reported in Table~\ref{tab:main_multi} in Appendix~\ref{sec:app_multistep}.
Seq-Drifting$^{S}$ produces competitive unconditional samples. On LM1B, it
outperforms the strongest four-step baseline and even beat several $1024$-step baselines, which directly reflect the effectiveness of our drifting framework.
Seq-Drifting$^{P}$ further reduces Gen-PPL relative to FMLM by $69.8\%$ on LM1B and $24.2\%$ on OpenWebText. 
%They both outperform the strongest four-step distilled model on LM1B while remaining competitive on OpenWebText. 
The entropy and Self-BLEU results in
Appendix~\ref{sec:app_selfbleu} show that they preserve generation diversity.
%Seq-Drifting$^{P}$ has higher entropy than Seq-Drifting$^{S}$. This may reflect differences in the token alternatives provided by the two support models.
To verify that the comparison is not tied to the GPT-2 Large evaluator, Appendix~\ref{sec:app_alt_eval}
recomputes Gen-PPL with Pythia-1B~\citep{biderman2023pythia} and OPT-1.3B~\citep{zhang2022opt}. The resulting comparisons remain consistent with the main table, showing that the observed
gains are not specific to the evaluator.
% Conditional generation (compact wrap table, sized like the unconditional distilled table).
% 64-token prefix; LM1B 64->64, OWT 64->512 (GPT-2 support model <= 1024 pos).
% Baselines = few-step distilled models, prefix-clamped infilling at 1/2/4 steps.
% PPL = conditional Gen-PPL under GPT-2 Large; Ent = sample unigram entropy.
% "--" = not in the results sheet (Duo+Di4C not run; several OWT/LM1B cells missing).
\begin{wraptable}{r}{0.46\textwidth}
%\vspace{0.45\baselineskip}
\centering
\scriptsize
\setlength{\tabcolsep}{2.3pt}
\renewcommand{\arraystretch}{0.84}
\caption{\textbf{Conditional generation.} A $64$-token prefix followed by a continuation.}
\label{tab:cond}
\begin{tabular}{l|c|cc|cc}
\toprule
\multirow{2}{*}{Method} & \multirow{2}{*}{Step} &
\multicolumn{2}{c|}{$64{\to}64$} & \multicolumn{2}{c}{$64{\to}512$} \\
\cmidrule(lr){3-4}\cmidrule(lr){5-6}
 & & Gen-PPL$\downarrow$ & Ent.\ & Gen-PPL$\downarrow$ & Ent.\ \\
\midrule
\multirow{3}{*}{FMLM}
 & 1 & 162.20 & 3.78 & 418.37 & 3.32 \\
 & 2 & 143.62 & 3.82 & 441.09 & 3.55 \\
 & 4 & 124.74 & 3.82 & 320.22 & 3.52 \\
\cmidrule(lr){1-6}
\multirow{3}{*}{Duo+DCD}
 & 1 & 972.09 & 3.77 & 7639.94 & 3.85 \\
 & 2 & 384.35 & 3.72 & 846.20 & 3.72 \\
 & 4 & 209.32 & 3.74 & 205.79 & 3.75 \\
\cmidrule(lr){1-6}
\multirow{3}{*}{Duo+Di4C}
 & 1 & 409.44 & 3.02 & 2784.97 & 3.74 \\
 & 2 & 507.34 & 3.30 & 1007.87 & 3.75 \\
 & 4 & 476.22 & 3.48 & 254.55 & 3.75 \\
\cmidrule(lr){1-6}
\multirow{3}{*}{MDLM+SDTT}
 & 1 & 912.05 & 3.75 & 2770.06 & 3.74 \\
 & 2 & 333.40 & 3.74 & 1077.27 & 3.75 \\
 & 4 & 139.73 & 3.74 & 298.50  & 3.75 \\
\cmidrule(lr){1-6}
\multirow{3}{*}{MDLM+Di4C}
 & 1 & 1018.23 & 3.83 & 2791.44 & 3.74 \\
 & 2 & 509.84  & 3.83 & 944.53  & 3.75 \\
 & 4 & 231.36  & 3.82 & 226.34  & 3.75 \\
\midrule
\rowcolor{gray!12}
\textbf{Seq-Drifting$^{S}$} & \textbf{1} & \textbf{79.13} & \textbf{3.74} & \textbf{119.13} & \textbf{4.10} \\
\rowcolor{gray!12}
\textbf{Seq-Drifting$^{P}$} & \textbf{1} & \textbf{76.16} & \textbf{4.40} & \textbf{168.34} & \textbf{4.78} \\
\bottomrule
\end{tabular}
\vspace{-35pt}
\end{wraptable}

The running-time in Figure~\ref{fig:main}(right) shows that Seq-Drifting has the lowest single-sample latency compared with the baselines.
%, at approximately $0.006$ seconds on LM1B and $0.010$ seconds on OpenWebText. 
Relative to four-step FMLM, which reaches a comparable quality level,
Seq-Drifting is $11.2\times$ faster on LM1B and $10.1\times$ faster on OpenWebText. A broader comparison
of sampling time across multi-step baselines is provided in Appendix~\ref{sec:app_timing}.
Qualitative samples are shown in Appendix~\ref{sec:app_case}.
%\Needspace{24\baselineskip}

\subsection{Conditional generation}
\label{sec:exp_cond}
Table~\ref{tab:cond} summarizes two conditional-generation settings, each with a 64-token context. On LM1B the model generates 64 output tokens. On OpenWebText it generates 512 output tokens so that the context and output remain within GPT-2's 1024-token context window. Both Seq-Drifting$^{S}$ and Seq-Drifting$^{P}$
outperform all few-step distilled baselines in both settings. Compared with unconditional
generation, the gap between Seq-Drifting$^{S}$ and Seq-Drifting$^{P}$ narrows, and Seq-Drifting$^{S}$ is even stronger on the $512$-token
output setting. One possible explanation is that the observed prefix narrows the plausible continuations,
reducing ambiguity in the support sets and making the conditional map easier to learn. Overall,
the results suggest that Seq-Drifting can preserve both fluency and conditional relevance in conditional generation.

\subsection{Ablations}
\label{sec:exp_ablation}
We report ablations in both the unconditional ($128$) and conditional ($64{\to}64$)
settings in Table~\ref{tab:ablation}. 
%Each ablation isolates one component of the drift construction. The results clarify how geometry, positive selection, repulsion, and warm-up contribute to the quality and stability of one-step generation.
The geometry ablation shows that the spherical update is essential. Replacing it with a
Euclidean update allows repulsion to increase the embedding norm and causes training to diverge.
The positive-selection ablation highlights that support construction should match the task. For
open-ended generation, fixed corpus-based targets do not follow the generator's evolving
contexts. When the attraction becomes sharp, this mismatch can produce oscillatory training and
rising generative perplexity and entropy (Appendix~\ref{sec:app_offpolicy}). For tasks with
explicit target answers, dataset-derived supervision can remain appropriate, as in our reasoning
experiments in \S\ref{sec:exp_downstream}. For repulsion, the cross-sample term is
not enough on its own. An intra-sequence term that pushes
each position off the other positions of the same generation is also necessary, and dropping it
(\emph{w/o intra}) collapses the output into within-sentence token repetition. In the conditional
setting, whether the cross-sample repulsion is restricted to samples that share a prefix makes little
difference. Finally, unconditional warm-up stabilizes
from-scratch conditional training by allowing the generator to learn fluency before it must also
model prefix dependence.
% Ablations: one design decision changed at a time; top row = full method (shown once).
% Uncond = length 128, Cond = 64->64. Cells are Gen-PPL (lower better) / unigram Entropy.
% Per-context repulsion and warm-up are conditional-only (no unconditional counterpart).
% \TODO{fill numbers from the ablation runs.}
\begin{table}[t]
\centering
\captionsetup{skip=5pt}
\caption{\textbf{Ablation study} on unconditional ($128$) and conditional ($64{\to}64$) generation.}
\label{tab:ablation}
\footnotesize
\setlength{\tabcolsep}{4pt}
\renewcommand{\arraystretch}{1.0}
\begin{tabular}{l l cc cc}
\toprule
 & & \multicolumn{2}{c}{Unconditional ($128$)} & \multicolumn{2}{c}{Conditional ($64{\to}64$)} \\
\cmidrule(lr){3-4}\cmidrule(lr){5-6}
Decision & Variant & Gen-PPL$\downarrow$ & Entropy & Gen-PPL$\downarrow$ & Entropy \\
\midrule
\multicolumn{2}{l}{\textbf{Seq-Drifting$^{P}$}} & 36.01 & 4.42 & 76.16 & 4.40 \\
\midrule
Geometry          & Euclidean             & \multicolumn{4}{c}{\emph{diverges (loss blow-up)}} \\
Positives         & off-policy            & $>$1000 & $>$6 & $>$10000 & $>$4 \\
Anti-repetition   & w/o intra             & 10.01 & 1.82 & 12.18 & 1.31 \\
Repulsion         & per-context (block)   & -- & -- & 81.27 & 4.27 \\
Warm-up           & w/o warm-up           & -- & -- & 53.57 & 3.87 \\
\bottomrule
\end{tabular}
\par\vspace{-5pt}
\end{table}

\begin{figure}[t]\centering
\includegraphics[width=\linewidth]{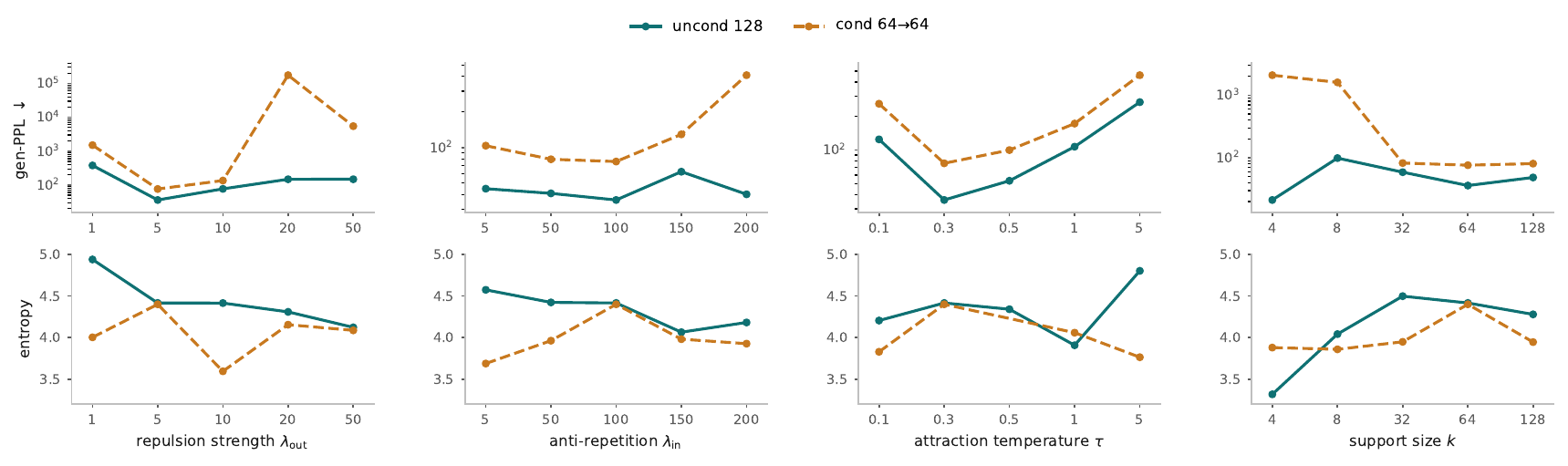}
\caption{\textbf{Hyperparameter sweeps} of the repulsion strength $\lambda_{\mathrm{out}}$, the anti-repetition weight
$\lambda_{\mathrm{in}}$, the attraction temperature $\tau$, and the support size $k$. Setting: unconditional $128$-token generation (solid), conditional $64{\to}64$ generation (dashed).}
\label{fig:hyper}
\par\vspace{-8pt}
\end{figure}
\subsection{Hyperparameters}
\label{sec:exp_hyper}
Figure~\ref{fig:hyper} further sweeps parameters $\lambda_{\mathrm{out}}, \lambda_{\mathrm{in}}, \tau$, and $k$. 
%on the same settings, with $\lambda_{\mathrm{out}}\in\{1,5,10,20,50\}$, $\lambda_{\mathrm{in}}\in\{5,50,100,150,200\}$, $\tau\in\{0.1,0.3,0.5,1,5\}$, and $k\in\{4,8,32,64,128\}$
We observe that the
first three parameters show the same broad pattern. Gen-PPL first improves and then worsens as the
parameter moves away from a favorable middle range, revealing a balance between a target that is too
weak and one that is too sharp or too disruptive.

$\lambda_{\mathrm{out}}$ has a larger effect than $\lambda_{\mathrm{in}}$. When repulsion between
different noise samples is too weak, the generator is encouraged to learn an average token direction
rather than preserve distinct modes. $\lambda_{\mathrm{in}}$ mainly controls repetition within a
sequence and therefore provides a narrower correction. For $\tau$, making the temperature too small
makes the target at each position nearly unique, discarding the useful ambiguity in the support tokens. For the support size, $k=4$ collapses the output (entropy below $3.5$). Increasing the support size
generally restores diversity and improves Gen-PPL.

\Needspace{6\baselineskip}
\subsection{Downstream task evaluation}
\label{sec:exp_downstream}
\paragraph{Translation and summarization.}
Following \citet{hu2026elf}, we evaluate Seq-Drifting on WMT14 (German--English
translation)~\citep{bojar2014findings} and XSum (abstractive summarization)~\citep{narayan2018don}.
Experimental details are provided in Appendix~\ref{sec:app_downstream_training}.
Table~\ref{tab:downstream} shows that Seq-Drifting outperforms the few-step baselines on translation and on most summarization metrics,
although a gap remains relative to the multi-step results.
We hypothesize that this gap partly reflects a mismatch between the
support-set candidates used during training and the single reference used
for evaluation. When evaluated against results sampled from the
support model, Seq-Drifting obtains substantially higher scores.
Scaling the support model further improves one-step generation
quality (Figure~\ref{fig:support_scaling}). This observation motivates
future work on using increasingly capable language models to construct
support sets, with the broader goal of turning existing LLMs into
efficient one-step generators.
% Shared float keeps the downstream table and support-scaling plot together.
\begin{table}[t]
\centering
\newsavebox{\downstreamresultbox}
\newsavebox{\downstreamfitbox}
\sbox{\downstreamresultbox}{%
\scriptsize
\setlength{\tabcolsep}{2pt}
\renewcommand{\arraystretch}{1.3}
\newcommand{\se}[1]{{\tiny $\pm #1$}}
\newcommand{\elfres}[1]{#1$^{\dagger}$}
\begin{tabular}{l|c|c|ccc}
\toprule
\multirow{2}{*}{Model} & \multirow{2}{*}{Step} &
\multicolumn{1}{c|}{WMT14} & \multicolumn{3}{c}{XSum} \\
 & & BLEU$\uparrow$ & ROUGE-1$\uparrow$ & ROUGE-2$\uparrow$ & ROUGE-L$\uparrow$ \\
\midrule
AR & {--} & \elfres{25.2} & \elfres{30.5\se{0.13}} & \elfres{10.2\se{0.11}} & \elfres{24.4\se{0.12}} \\
\cmidrule(lr){1-6}
\multirow{2}{*}{MDLM}
 & 4 & $<$ 10 & 19.0\se{0.06} & 2.4\se{0.01} & 13.2\se{0.04}  \\
 & 64 & \elfres{18.4} & \elfres{33.4\se{0.11}} & \elfres{11.6\se{0.10}} & \elfres{25.8\se{0.10}} \\
\cmidrule(lr){1-6}
\multirow{2}{*}{Duo}
 & 4 & $<$ 10 & 24.4\se{0.06} & 5.6\se{0.01} & 18.1\se{0.04}  \\
 & 1000 & \elfres{21.3} & \elfres{31.4\se{0.12}} & \elfres{10.1\se{0.10}} & \elfres{25.0\se{0.12}} \\
\cmidrule(lr){1-6}
\multirow{2}{*}{ELF}
 & 1 & 11.5 & 23.7\se{0.11} & 5.7\se{0.08} & 19.4\se{0.10} \\
 & 64 & \elfres{26.4} & \elfres{36.0\se{0.13}} & \elfres{12.2\se{0.11}} & \elfres{27.8\se{0.12}} \\
\midrule
\textbf{Seq-Drifting (vs.\ dataset)} & 1 & 18.6 & 24.7\se{0.36} & 5.3\se{0.24} & 20.1\se{0.32} \\
\textbf{Seq-Drifting (vs.\ AR)} & 1 & 30.8 & 33.2\se{0.44} & 11.6\se{0.41} & 28.7\se{0.44} \\
\bottomrule
\end{tabular}%
}
\sbox{\downstreamfitbox}{\resizebox{0.68\linewidth}{!}{\usebox{\downstreamresultbox}}}
% The plot canvas and table have exactly the same height.
\begin{minipage}[t]{0.68\linewidth}
\vspace{0pt}\centering
\usebox{\downstreamfitbox}
\end{minipage}\hfill
\begin{minipage}[t]{0.30\linewidth}
\vspace{0pt}\centering
\includegraphics[width=\linewidth,height=\dimexpr\ht\downstreamfitbox+\dp\downstreamfitbox\relax]{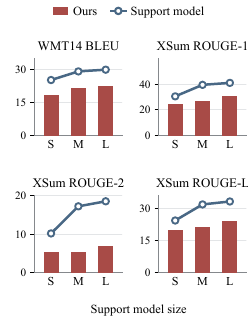}
\end{minipage}
\par\vspace{2pt}
\begin{minipage}[t]{0.68\linewidth}
\captionsetup{font=small,skip=3pt}
\caption{Performance on WMT14 De--En and XSum. The two Seq-Drifting rows evaluate results on the dataset ground truth and
samples from our support model, respectively. $\dagger$ denotes results taken from~\cite{hu2026elf}.}
\label{tab:downstream}
\end{minipage}\hfill
\begin{minipage}[t]{0.30\linewidth}
\captionsetup{font=small,skip=3pt,justification=raggedright,singlelinecheck=false}
\captionof{figure}{Support model scaling. S, M, and L denote small, medium, and large GPT-2.}
\label{fig:support_scaling}
\end{minipage}
\end{table}

\paragraph{Reasoning.}
We further evaluate Seq-Drifting on math reasoning (SVAMP~\citep{patel2021nlp} and ASDiv~\citep{miao2020diverse}) and logical reasoning
(ProofWriter~\citep{tafjord2021proofwriter}). We scale Seq-Drifting and all the baselines to $0.5$B. The implementation and training details are provided in Appendix~\ref{sec:app_reasoning}.
Despite using a single forward pass, Seq-Drifting approaches the accuracy of existing
latent-reasoning method Coconut~\citep{hao2025training} on the mathematical benchmarks and remains competitive with AR on ProofWriter, while achieving substantially lower per-sample latency.
These results show that a drifting field can also be learned from supervision with explicit target
answers, extending the one-step framework beyond open-ended generation.
\begin{table}[t]
\centering
\caption{Performance on SVAMP, ASDiv, and ProofWriter.}
\label{tab:reasoning}
\scriptsize
\setlength{\tabcolsep}{5pt}
\renewcommand{\arraystretch}{1.15}
\newcommand{\se}[1]{{\tiny $\pm #1$}}
\begin{tabular}{l|cc|cc|cc}
\toprule
\multirow{2}{*}{Model} & \multicolumn{2}{c|}{SVAMP} &
\multicolumn{2}{c|}{ASDiv} & \multicolumn{2}{c}{ProofWriter} \\
\cmidrule(lr){2-3}\cmidrule(lr){4-5}\cmidrule(lr){6-7}
 & ACC$\uparrow$ & Time/sample (ms)$\downarrow$ & ACC$\uparrow$ & Time/sample (ms)$\downarrow$ & ACC$\uparrow$ & Time/sample (ms)$\downarrow$ \\
\midrule
%AR-MAWPS & 27.92\se{0.36} & 49.42\se{4.36} & 46.58\se{0.50} & 43.17\se{0.40} & 85.90\se{0.38} & 41.90\se{0.10} \\
AR & 27.92\se{0.36} & 49.42\se{4.36} & 44.97\se{0.50} & 31.07\se{1.18} & 85.90\se{0.38} & 41.90\se{0.10} \\
%NAR-MAWPS & 4.72\se{0.76} & 2.70\se{0.04} & 1.83\se{0.48} & 2.60\se{0.05} & 0.00\se{0.00} & 1.42\se{0.06} \\
NAR & 4.72\se{0.76} & 2.70\se{0.04} & 1.73\se{0.57} & 1.30\se{0.07} & 0.00\se{0.00} & 1.42\se{0.06} \\
%iCoT-SI-MAWPS & 18.46\se{0.54} & 4.18\se{0.02} & 21.33\se{0.43} & 7.73\se{0.01} & -- & -- \\
Coconut & 17.76\se{0.01} & 25.80\se{0.52} & 22.59\se{0.01} & 26.63\se{0.13} & -- & -- \\
\midrule
\rowcolor{gray!12}
%\textbf{Seq-Drifting}-MAWPS & 13.60\se{0.50} & 5.88\se{0.17} & 13.39\se{0.20} & 5.43\se{0.12} & 79.31\se{0.47} & 3.68\se{0.02} \\
\textbf{Seq-Drifting} & 13.60\se{0.50} & 5.88\se{0.17} & 19.11\se{0.79} & 1.05\se{0.03} & 79.31\se{0.47} & 3.68\se{0.02} \\
\bottomrule
\end{tabular}
\par\vspace{-8pt}
\end{table}

% =====================================================================
\section{Conclusion}
In this paper, we presented Seq-Drifting, a framework for directly learning one-step text generation by drifting
sequences of continuous token representations. Context-dependent support sets provide multiple
plausible targets at each position, while mode-seeking attraction, repulsion, and spherical
geometry help preserve coherent and diverse outputs. Across unconditional generation, conditional continuation, translation, summarization, and reasoning, Seq-Drifting provides a competitive one-step operating point and outperforms the evaluated few-step baselines at a fraction of their inference
cost. The remaining gap to the strongest multi-step systems motivates future work on stronger
support models, richer semantic supervision, and improved token recovery.

% =====================================================================
\subsection*{AI Use Statement}
In this work, we used generative AI tools to edit the manuscript for readability and clarity. All AI-assisted text was reviewed and revised by the authors. The authors take full responsibility for the final content of this work, including all claims, analyses, and conclusions.

% \subsection*{Ethics Statement}
% This work does not involve human subjects, private or sensitive data, or applications that raise additional ethical concerns. We have reviewed the ICLR Code of Ethics and certify that this submission complies with it.

% \subsection*{Reproducibility Statement}
% We describe the model architecture, training procedure, support-set construction, drift objective, datasets, evaluation protocols, and hyperparameters in the main paper and appendix. We provide an anonymized code repository at \url{https://github.com/jyliu-ai/Seq-Drifting} and include it with the supplementary materials. The repository contains the implementation and instructions needed to reproduce the reported experiments.

\bibliography{iclr2027_conference}
\bibliographystyle{iclr2027_conference}

\appendix
\section{Method details and analysis}

\subsection{Generator architecture}
\label{sec:app_generator}
A shared noise vector $\epsilon\in\mathbb{R}^{C}$ is projected to $T\!\cdot\!D$ dimensions,
reshaped to $(T,D)$, and combined with learned positional embeddings. Our generator $G_\theta$ is a $12$-layer pre-LN Transformer with width $768$, $12$ attention heads, and an
FFN width of $3072$. The noise dimension is $C=128$. We use QK-norm on the per-head queries and keys
and a LayerScale gate on each residual branch as stabilization mechanisms. For conditional
generation, conditioning token embeddings are supplied as a prefix and the output positions
access them through ungated cross-attention. 

%\subsection{Support-model training}
%\label{sec:app_support_training}
%For \textbf{Seq-Drifting$^{S}$}, the support model is a standard GPT-2-small Transformer with
%124M parameters, 12 layers, width $768$, 12 attention heads, and FFN width $3072$. We train it
%from scratch on the corresponding corpus with the causal next-token objective for $10^{5}$ update
%steps and then freeze its parameters. During Seq-Drifting training, its next-token predictions
%under each generated context define the position-wise support sets. The Seq-Drifting generator is
%trained for $2\times10^{5}$ steps. For \textbf{Seq-Drifting$^{P}$}, we use the corresponding public
%GPT-2 checkpoint and perform no additional support-model training. The support model is queried
%only during training and is not used for one-step inference.
\subsection{A conditional fixed-point analysis}
\label{sec:app_fixedpoint}
We analyze conditions under which matching the generated and data distributions makes the idealized drift field zero.
For $q_\theta=p_{\mathrm{data}}$ to be a fixed point of training, the drift $V$, and hence the loss of
Eq.~(\ref{eq:drift_loss}), must be zero there. Anti-symmetry $V_{p,q}=-V_{q,p}$ is one way to guarantee
this and is how the plain mean-shift field of \citet{deng2026generativemodelingdrifting} does it, but it
is a \emph{sufficient} condition, not a necessary one. Seq-Drifting's per-position field is not
anti-symmetric, so instead of appealing to that argument we check $V=\boldsymbol{0}$ at
$q_\theta=p_{\mathrm{data}}$ directly.

\paragraph{The field.} At position $t$, with unit embedding
$\boldsymbol{e}_t$, the drift of \S\ref{sec:method} is
\begin{equation}
\boldsymbol{F}_t=\underbrace{\sum_{\boldsymbol{w}\in\mathcal{S}_t}\varphi(\boldsymbol{w},\boldsymbol{e}_t)(\boldsymbol{w}-\boldsymbol{e}_t)}_{\text{attraction }\boldsymbol{u}_t}
-\lambda_{\mathrm{out}}\!\!\sum_{\boldsymbol{n}\in N_t^{\mathrm{out}}}\!\!\varphi(\boldsymbol{n},\boldsymbol{e}_t)(\boldsymbol{n}-\boldsymbol{e}_t)
-\lambda_{\mathrm{in}}\!\!\sum_{\boldsymbol{n}\in N_t^{\mathrm{in}}}\!\!\varphi(\boldsymbol{n},\boldsymbol{e}_t)(\boldsymbol{n}-\boldsymbol{e}_t),
\label{eq:app_sdfield}
\end{equation}
with $\varphi(\boldsymbol{a},\boldsymbol{b})\propto\exp\!\big(\!-d(\boldsymbol{a},\boldsymbol{b})/(s\,\tau)\big)$
and $d(\boldsymbol{a},\boldsymbol{b})=1-\boldsymbol{a}^{\top}\boldsymbol{b}$ the angular kernel, normalized
over each set. The attraction pulls $\boldsymbol{e}_t$ toward the support set $\mathcal{S}_t$ of plausible
tokens, and the two repulsions push it off the other generations $N_t^{\mathrm{out}}$ and the other
positions of its own sequence $N_t^{\mathrm{in}}$.

\paragraph{Attraction and rescaled repulsion at matching distributions.} At $q_\theta=p_{\mathrm{data}}$ the generator reproduces the
data. Suppose each output embedding coincides with a real token embedding, $\boldsymbol{e}_t=\boldsymbol{w}^{\star}$
with $\boldsymbol{w}^{\star}\in\mathcal{S}_t$. Importantly, this does not depend on any particular support model. In the mode-seeking limit $\tau\!\to\!0$, a generation already sitting on a supported token has
nothing to be pulled toward, so the attraction is zero,
\begin{equation}
\boldsymbol{u}_t=\sum_{\boldsymbol{w}\in\mathcal{S}_t}\varphi(\boldsymbol{w},\boldsymbol{e}_t)(\boldsymbol{w}-\boldsymbol{e}_t)\;\longrightarrow\;\varphi(\boldsymbol{w}^{\star},\boldsymbol{e}_t)(\boldsymbol{w}^{\star}-\boldsymbol{e}_t)=\boldsymbol{0},
\end{equation}
%exactly in the mode-seeking limit $\tau\!\to\!0$, where $\varphi$ concentrates on the zero-distance candidate $\boldsymbol{w}^{\star}=\boldsymbol{e}_t$, and to good approximation at the small $\tau$ we use.
The repulsions then vanish with it, because each is rescaled to the attraction magnitude before entering
$\boldsymbol{F}_t$, with coefficient $a_{\mathrm{rms}}/r_{\mathrm{rms}}$ and
$a_{\mathrm{rms}}=\mathrm{RMS}(\boldsymbol{u}_t)$, so the repulsion magnitude is $\propto a_{\mathrm{rms}}\to0$.
Under these conditions, $\boldsymbol{F}_t$ is zero at every position and $q_\theta=p_{\mathrm{data}}$ is a fixed point
of the idealized field. The attraction-relative rescaling is what lets a strong anti-collapse repulsion
coexist with this fixed point, because the repulsion is rescaled by the attraction magnitude and
therefore switches off with it.

%\paragraph{Scope of the argument.}
%The calculation above uses the zero-temperature mode-seeking limit to make the nearest supported candidate explicit.
%In practice, the temperature is positive, so other supported candidates can still contribute to the attraction. When
%their distances are sufficiently larger than that of the nearest candidate relative to the kernel temperature scale,
%their weights are exponentially suppressed. The resulting perturbation can therefore be small, while the exact
%zero-drift equality is a property of the idealized calculation.

In practice, the temperature is positive and the decoded token distribution can match the data distribution even when
generated representations slightly differ from their corresponding vocabulary
embeddings. We now quantify how such discrepancies
affect the drift. Let $\boldsymbol{w}_t^\star=\boldsymbol{w}_{\mathrm{tok}_t}$ and consider generated
sequences for which $\boldsymbol{w}_t^\star\in\mathcal{S}_t$ is the unique nearest
candidate. Since the kernel weights sum to one, the attraction decomposes exactly as
\begin{equation}
\boldsymbol{u}_t=\boldsymbol{w}_t^\star-\boldsymbol{e}_t+
\sum_{\boldsymbol{w}\in\mathcal{S}_t\setminus\{\boldsymbol{w}_t^\star\}}
\varphi(\boldsymbol{w},\boldsymbol{e}_t)(\boldsymbol{w}-\boldsymbol{w}_t^\star).
\label{eq:app_attraction_decomposition}
\end{equation}
The first term is the representation discrepancy, and the second is the finite-temperature
correction. For a support set with at least two candidates, define
\begin{equation}
\begin{aligned}
\Delta_t&=\min_{\boldsymbol{w}\in\mathcal{S}_t\setminus\{\boldsymbol{w}_t^\star\}}
\big[d(\boldsymbol{e}_t,\boldsymbol{w})-d(\boldsymbol{e}_t,\boldsymbol{w}_t^\star)\big]>0,\\
b_t&=(|\mathcal{S}_t|-1)\exp\!\left(-\frac{\Delta_t}{s_t\tau}\right).
\end{aligned}
\end{equation}
Each other candidate has at most
$\exp(-\Delta_t/(s_t\tau))$ times the nearest candidate's kernel weight, giving
$1-\varphi(\boldsymbol{w}_t^\star,\boldsymbol{e}_t)\leq b_t/(1+b_t)$.
All vocabulary embeddings have unit norm, so their pairwise distances are at most two.
It follows that
\begin{equation}
\big\|\boldsymbol{u}_t-(\boldsymbol{w}_t^\star-\boldsymbol{e}_t)\big\|_2
\leq2\big[1-\varphi(\boldsymbol{w}_t^\star,\boldsymbol{e}_t)\big]
\leq\frac{2b_t}{1+b_t}.
\label{eq:app_temperature_correction}
\end{equation}
For singleton supports, the sole candidate has weight one and we set $b_t=0$.
Define $\|\boldsymbol{a}\|_{\mathrm{RMS}}=(\mathbb{E}\|\boldsymbol{a}_t\|_2^2)^{1/2}$,
averaging over generated samples and positions, and write
\begin{equation}
\delta=\big(\mathbb{E}\|\boldsymbol{e}_t-\boldsymbol{w}_t^\star\|_2^2\big)^{1/2},
\qquad
\rho_\tau=\left[\mathbb{E}\left(\frac{2b_t}{1+b_t}\right)^2\right]^{1/2}.
\end{equation}
The $L^2$ triangle inequality gives $\|\boldsymbol{u}\|_{\mathrm{RMS}}\leq\delta+\rho_\tau$.
Under the attraction-relative rescaling above, each rescaled repulsion has RMS no greater
than that of the attraction, using the same normalization groups and leaving zero repulsion
at zero. Tangent projection cannot increase the force norm, and the geodesic displacement
$V_t=goal_t-\boldsymbol{e}_t$ satisfies
$\|V_t\|_2=2|\sin(\alpha_t/2)|\leq\alpha_t\leq\|\boldsymbol{F}^{\prime}_t\|_2$,
where $\alpha_t$ is the step angle at position $t$ and zero tangent force gives zero
displacement. Applying the triangle inequality to the combined force therefore yields
\begin{equation}
\|V\|_{\mathrm{RMS}}\leq\|\boldsymbol{F}\|_{\mathrm{RMS}}
\leq(1+\lambda_{\mathrm{out}}+\lambda_{\mathrm{in}})(\delta+\rho_\tau).
\label{eq:app_drift_bound}
\end{equation}
The loss averaged over samples and positions is bounded by the square of this expression.
For a fixed distribution of representations satisfying the support and uniqueness conditions
above, $\rho_\tau\to0$ as $\tau\to0$ by dominated convergence, since $b_t\to0$ and
$2b_t/(1+b_t)\leq2$. At positive temperature, the bound remains small when the representation
discrepancy and the temperature correction are small enough for the chosen repulsion weights.
Thus, when continuous
representations do not coincide exactly, matching the decoded distribution can ensure that the drift is bounded.

\subsection{Seq-Drifting algorithm}
\label{sec:app_algorithm}
The pseudocode in Figure~\ref{alg:seqdrifting} summarizes Seq-Drifting training with on-policy support sets
constructed by a frozen support model. Each iteration generates token representations,
builds context-dependent support sets, and combines attraction and repulsion to form
supervision targets on the sphere. The generator is then updated by regressing toward
these fixed targets, with optional prefixes enabling conditional generation.

\begin{figure}[t]
\centering
\begin{minipage}{0.94\linewidth}
\hrule\vspace{3pt}
\noindent\textbf{Algorithm 1: Seq-Drifting training.}\vspace{2pt}
\hrule\vspace{2pt}
\small
\textbf{Input:} generator $G_\theta$, frozen support model $p_{\psi}$, normalized token table
$\{\boldsymbol{w}_v\}_{v\in\mathcal{V}}$, batch size $B$, support size $k$,
$\tau,\lambda_{\mathrm{out}},\lambda_{\mathrm{in}},\alpha_{\max}$.\\
\textbf{repeat}\\
\hspace*{1em}Sample noise $\epsilon_1,\ldots,\epsilon_B$ and optional prefixes
$\boldsymbol{p}_1,\ldots,\boldsymbol{p}_B$.\\
\hspace*{1em}Compute one-step outputs
$\boldsymbol{h}_{b,1:T}=G_\theta(\epsilon_b,\boldsymbol{p}_b)$ and normalize
$\boldsymbol{e}_{b,t}=\boldsymbol{h}_{b,t}/\|\boldsymbol{h}_{b,t}\|$.\\
\hspace*{1em}Decode provisional tokens
$\hat{x}_{b,t}=\arg\max_v \boldsymbol{e}_{b,t}^{\top}\boldsymbol{w}_v$.\\
\hspace*{1em}Run one causal forward pass of $p_\psi$ on each generated context and form
$\mathcal{S}_{b,t}$ from the top-$k$ plausible next tokens.\\
\hspace*{1em}\textbf{for} each sample $b$ and position $t$ \textbf{do}\\
\hspace*{2em}Compute angular attraction $\boldsymbol{u}_{b,t}$ from $\mathcal{S}_{b,t}$.\\
\hspace*{2em}Compute cross-sample and intra-sequence repulsions
$\boldsymbol{r}^{\mathrm{out}}_{b,t}$ and $\boldsymbol{r}^{\mathrm{in}}_{b,t}$.\\
\hspace*{2em}Rescale each repulsion to the attraction magnitude:
$\widetilde{\boldsymbol{r}}^{j}_{b,t}=(a_{\mathrm{rms}}/r_{\mathrm{rms}}^{j})\boldsymbol{r}^{j}_{b,t}$
for $j\in\{\mathrm{out},\mathrm{in}\}$, where the RMS values are computed over the batch
and positions; set $\widetilde{\boldsymbol{r}}^{j}=\boldsymbol{0}$ when $r_{\mathrm{rms}}^{j}=0$.\\
\hspace*{2em}Set
$\boldsymbol{F}_{b,t}=\boldsymbol{u}_{b,t}
+\lambda_{\mathrm{out}}\widetilde{\boldsymbol{r}}^{\mathrm{out}}_{b,t}
+\lambda_{\mathrm{in}}\widetilde{\boldsymbol{r}}^{\mathrm{in}}_{b,t}$.\\
\hspace*{2em}Project $\boldsymbol{F}_{b,t}$ to the tangent plane and map it to
$goal_{b,t}$ by Eq.~(\ref{eq:geodesic}).\\
\hspace*{1em}\textbf{end for}\\
\hspace*{1em}Update $\theta$ with
$\nabla_\theta \frac{1}{BT}\sum_{b,t}\|\boldsymbol{e}_{b,t}
-\operatorname{sg}(goal_{b,t})\|_2^2$.\\
\textbf{until} convergence
%\\
%\textbf{Inference:} output
%$\arg\max_v \boldsymbol{e}_{t}^{\top}\boldsymbol{w}_v$ from one call to $G_\theta$.
\vspace{3pt}\hrule
\end{minipage}
\caption{Seq-Drifting training pseudocode.}
\label{alg:seqdrifting}
\end{figure}

\section{Additional results}

\subsection{Self-BLEU}
\label{sec:app_selfbleu}
Table~\ref{tab:selfbleu} reports Self-BLEU~\citep{zhu2018texygen} for the few-step distilled baselines,
at each of their step counts ($1$, $2$, $4$), and for Seq-Drifting at a single step. Self-BLEU measures the
n-gram overlap among a model's own generations. A fully mode-collapsed generator that emits the same
sample every time attains the maximum of $1.0$.
% Appendix Self-BLEU for the few-step distilled baselines (at their 1/2/4 steps) and Seq-Drifting.
% "Mode collapse" reference row = 1.0. Reported for completeness, not as a ranking.
\begin{table}[t]
\centering
\caption{\textbf{Self-BLEU across the step budget.} Per-dataset Self-BLEU for the few-step distilled
models and Seq-Drifting; a fully mode-collapsed generator scores $1.0$ (bottom row).}
\label{tab:selfbleu}
\small
\setlength{\tabcolsep}{6pt}
\renewcommand{\arraystretch}{0.95}
\begin{tabular}{l|c|cc}
\toprule
\multirow{2}{*}{Method} & \multirow{2}{*}{Step} & \multicolumn{2}{c}{Self-BLEU} \\
\cmidrule(lr){3-4}
 & & LM1B & OpenWebText \\
\midrule
\multirow{3}{*}{FMLM}
 & 1 & 0.201 & 0.249 \\
 & 2 & 0.198 & 0.251 \\
 & 4 & 0.204 & 0.260 \\
\cmidrule(lr){1-4}
\multirow{3}{*}{Duo+DCD}
 & 1 & 0.042 & 0.016 \\
 & 2 & 0.065 & 0.064 \\
 & 4 & 0.093 & 0.140 \\
\cmidrule(lr){1-4}
\multirow{3}{*}{Duo+Di4C}
 & 1 & 0.054 & 0.082 \\
 & 2 & {--}  & 0.105 \\
 & 4 & {--}  & 0.190 \\
\cmidrule(lr){1-4}
\multirow{3}{*}{MDLM+SDTT}
 & 1 & 0.049 & 0.082 \\
 & 2 & 0.061 & 0.098 \\
 & 4 & 0.104 & 0.177 \\
\cmidrule(lr){1-4}
\multirow{3}{*}{MDLM+Di4C}
 & 1 & 0.023 & 0.080 \\
 & 2 & 0.039  & 0.111 \\
 & 4 & 0.061  & 0.204 \\
\midrule
\rowcolor{gray!12}
\textbf{Seq-Drifting$^{S}$} & \textbf{1} & \textbf{0.066} & \textbf{0.146} \\
\rowcolor{gray!12}
\textbf{Seq-Drifting$^{P}$} & \textbf{1} & \textbf{0.042} & \textbf{0.206} \\
\midrule
\emph{Mode collapse} & -- & 1.00 & 1.00 \\
\bottomrule
\end{tabular}
\end{table}

\subsection{Sampling time across step budgets}
\label{sec:app_timing}
\begin{figure}[t]
\centering
\includegraphics[width=\linewidth]{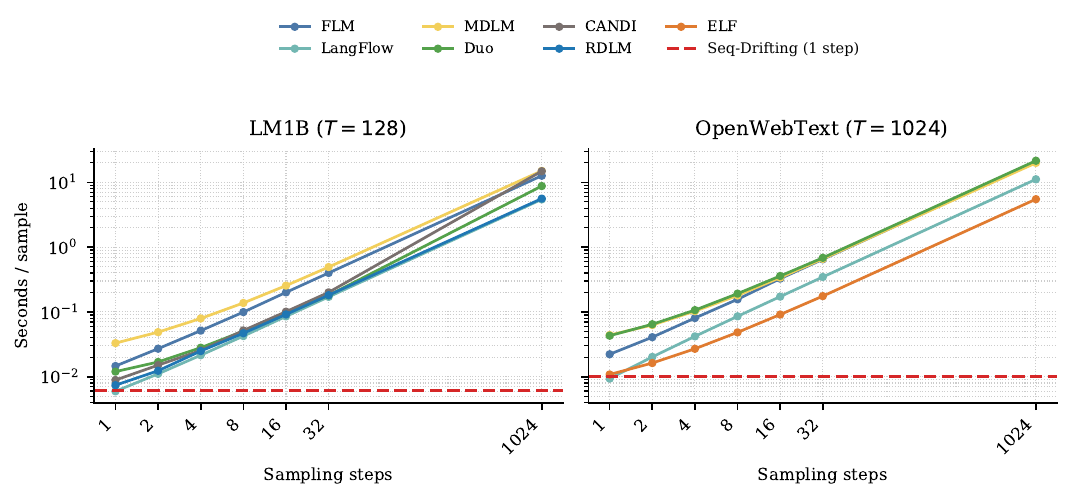}
\caption{\textbf{Sampling time across step budgets.} Seconds per sample on LM1B ($T=128$)
and OpenWebText ($T=1024$). The x-axis is logarithmic in the number of sampling steps and the
y-axis is logarithmic in seconds per sample. The dashed line marks Seq-Drifting.}
\label{fig:timing_steps}
\end{figure}

Figure~\ref{fig:timing_steps} reports seconds per sample as a function of the sampling budget for the multi-step baselines. The horizontal dashed line marks the one-step latency of Seq-Drifting. Sampling time grows with the number of denoising steps, while Seq-Drifting remains constant at one step.

\subsection{Multi-step baselines}
\label{sec:app_multistep}
Table~\ref{tab:main_multi} gives the full multi-step baselines (FLM, Duo, MDLM, CANDI, RDLM, LangFlow, ELF)
at their $1$/$8$/$1024$ step budgets on both datasets, with Seq-Drifting included for reference. Table~\ref{tab:cond_multi}
gives the multi-step baselines on the \emph{conditional} task at the same step budgets.

\paragraph{Multi-step baseline implementations.}
For FLM, we use the authors' released checkpoints for LM1B and OpenWebText
\citep{lee2026one}. For Duo, MDLM, and CANDI on LM1B, we use the
reproduction checkpoints distributed by \cite{lee2026one}. On OpenWebText,
we use the released Duo and MDLM weights
\citep{sahoo2024simple,sahoo2025diffusion}. For CANDI on OpenWebText, we
report the published result from \citet{lee2026one}, while no public
conditional OpenWebText checkpoint is available. For RDLM, we use the
released LM1B checkpoint and its accompanying SDE configuration
\citep{jo2026continuous}. We do not report OpenWebText results because no
corresponding public checkpoint is available. For LangFlow, we use the
authors' released checkpoints~\citep{chen2026langflow}. For ELF, we use the
released OpenWebText results and checkpoints~\citep{hu2026elf}; no
corresponding LM1B checkpoint is available. 
% Appendix: full multi-step baselines (moved out of the main text). Gen-PPL + Entropy per dataset.
% "--" = not run (OWT CANDI/RDLM incomplete).
\begin{table}[t]
\centering
\caption{\textbf{Unconditional generation: multi-step models} across the step budget, with Seq-Drifting at a single pass.
LM1B ($T{=}128$) and OpenWebText ($T{=}1024$). A dash indicates that no corresponding released checkpoint or published reference result was available for that dataset--model setting.}
\label{tab:main_multi}
\small
\setlength{\tabcolsep}{6pt}
\renewcommand{\arraystretch}{0.95}
\begin{tabular}{l|c|cc|cc}
\toprule
\multirow{2}{*}{Method} & \multirow{2}{*}{Step} &
\multicolumn{2}{c|}{LM1B \, ($T{=}128$)} & \multicolumn{2}{c}{OpenWebText \, ($T{=}1024$)} \\
\cmidrule(lr){3-4}\cmidrule(lr){5-6}
 & & Gen-PPL$\downarrow$ & Entropy & Gen-PPL$\downarrow$ & Entropy \\
\midrule
\multirow{3}{*}{FLM}
 & 1    & 6.27   & 1.15 & 27.54  & 1.53 \\
 & 8    & 215.84 & 4.24 & 453.49 & 5.20 \\
 & 1024 & 85.24  & 4.29 & 63.97  & 5.35 \\
\cmidrule(lr){1-6}
\multirow{3}{*}{Duo}
 & 1    & 1027.77 & 4.34 & 4007.98 & 5.80 \\
 & 8    & 121.37  & 4.27 & 196.45  & 5.57 \\
 & 1024 & 85.97   & 4.30 & 78.11   & 5.54 \\
\cmidrule(lr){1-6}
\multirow{3}{*}{MDLM}
 & 1    & 1348.21 & 4.37 & 4180.42 & 5.82 \\
 & 8    & 177.21  & 4.33 & 785.33  & 5.87 \\
 & 1024 & 69.39   & 4.30 & 42.36   & 5.30 \\
\cmidrule(lr){1-6}
CANDI & 1024 & 108.07 & 4.35 & 143.13 & 5.71 \\
\cmidrule(lr){1-6}
\multirow{3}{*}{RDLM}
 & 1    & 22.70  & 3.20 & {--} & {--} \\
 & 8    & 229.32 & 4.01 & {--} & {--} \\
 & 1024 & 274.94 & 4.35 & {--} & {--} \\
\cmidrule(lr){1-6}
\multirow{3}{*}{LangFlow}
 & 1    & 1.69 & 0.30 & 14.53 & 1.00 \\
 & 8    & 536.60 & 4.38 & 514.24 & 5.80 \\
 & 1024 & 121.73 & 4.31 & 36.53  & 5.25 \\
\cmidrule(lr){1-6}
\multirow{3}{*}{ELF}
 & 1    & {--} & {--} & 2.97  & 1.99 \\
 & 8    & {--} & {--} & 70.95 & 5.25 \\
 & 1024 & {--} & {--} & 7.73  & 4.74 \\
\midrule
\rowcolor{gray!12}
\textbf{Seq-Drifting$^{S}$} & \textbf{1} & \textbf{93.81} & \textbf{4.21} & \textbf{183.98} & \textbf{4.95} \\
\rowcolor{gray!12}
\textbf{Seq-Drifting$^{P}$} & \textbf{1} & \textbf{36.01} & \textbf{4.42} & \textbf{128.33} & \textbf{5.08} \\
\bottomrule
\end{tabular}
\end{table}

% Appendix: conditional multi-step baselines (moved out of the main text).
% 64-token prefix; 64->64 (LM1B) and 64->512 (OWT). Gen-PPL = conditional Gen-PPL under GPT-2 Large.
% "--" = not in the results sheet (some OWT / step budgets not run; CANDI few-step, RDLM many-step only).
\begin{table}[t]
\centering
\caption{\textbf{Conditional generation: multi-step models} across the step budget, with Seq-Drifting at a single
pass. A $64$-token prefix, then a continuation ($64{\to}64$ on LM1B, $64{\to}512$ on OWT). A dash indicates that no corresponding released checkpoint or published reference result was available for that dataset--model setting.}
\label{tab:cond_multi}
\small
\setlength{\tabcolsep}{6pt}
\renewcommand{\arraystretch}{0.95}
\begin{tabular}{l|c|cc|cc}
\toprule
\multirow{2}{*}{Method} & \multirow{2}{*}{Step} &
\multicolumn{2}{c|}{$64{\to}64$} & \multicolumn{2}{c}{$64{\to}512$} \\
\cmidrule(lr){3-4}\cmidrule(lr){5-6}
 & & Gen-PPL$\downarrow$ & Entropy & Gen-PPL$\downarrow$ & Entropy \\
\midrule
\multirow{3}{*}{FLM}
 & 1    & 1.95   & 0.83 & 201.21 & 2.18 \\
 & 8    & 6.00   & 1.81 & 522.85 & 3.63 \\
 & 1024 & 9.47   & 2.19 & 304.01 & 3.44 \\
\cmidrule(lr){1-6}
\multirow{3}{*}{Duo}
 & 1    & 1103.17 & 3.83 & 7616.55 & 3.86 \\
 & 8    & 99.98   & 3.78 & 181.91  & 3.81 \\
 & 1024 & 72.04   & 3.81 & 84.65   & 3.82 \\
\cmidrule(lr){1-6}
\multirow{3}{*}{MDLM}
 & 1    & 1064.26 & 3.83 & 6026.52 & 3.85 \\
 & 8    & 147.58  & 3.82 & 490.74  & 3.86 \\
 & 1024 & 58.64   & 3.80 & 61.71   & 3.79 \\
\cmidrule(lr){1-6}
CANDI & 1024 & 93.09 & 3.84 & {--} & {--} \\
\cmidrule(lr){1-6}
RDLM & 1024 & 72.79 & 3.21 & {--} & {--} \\
\cmidrule(lr){1-6}
\multirow{3}{*}{LangFlow}
 & 1    & 2.94   & 0.32 & 72.60    & 0.74 \\
 & 8    & 242.77 & 3.87 & 743.79  & 3.78 \\
 & 1024 & 64.05  & 3.81 & 83.30   & 3.31 \\
\cmidrule(lr){1-6}
\multirow{3}{*}{ELF}
 & 1    & {--} & {--} & 4.50  & 1.82 \\
 & 8    & {--} & {--} & 91.49 & 4.99 \\
 & 1024 & {--} & {--} & 10.59  & 4.49 \\
\midrule
\rowcolor{gray!12}
\textbf{Seq-Drifting$^{S}$} & \textbf{1} & \textbf{79.13} & \textbf{3.74} & \textbf{119.13} & \textbf{4.10} \\
\rowcolor{gray!12}
\textbf{Seq-Drifting$^{P}$} & \textbf{1} & \textbf{76.16} & \textbf{4.40} & \textbf{168.34} & \textbf{4.78} \\
\bottomrule
\end{tabular}
\end{table}

\subsection{Alternative Gen-PPL evaluators}
\label{sec:app_alt_eval}
The main comparison reports Gen-PPL under GPT-2 Large. To test whether the results depend on that
evaluator, we recompute unconditional Gen-PPL with Pythia-1B and OPT-1.3B in Table~\ref{tab:alt_eval}. All other sampling settings are unchanged.
% Unconditional Gen-PPL recomputed with alternative language-model evaluators.
\begin{table}[t]
\centering
\caption{\textbf{Unconditional Gen-PPL under alternative evaluators.} Gen-PPL computed with
Pythia-1B and OPT-1.3B for LM1B ($T{=}128$) and OpenWebText ($T{=}1024$).}
\label{tab:alt_eval}
\small
\setlength{\tabcolsep}{3.5pt}
\renewcommand{\arraystretch}{0.95}
\begin{tabular}{l|c|cc|cc}
\toprule
\multirow{2}{*}{Method} & \multirow{2}{*}{Step} &
\multicolumn{2}{c|}{LM1B} & \multicolumn{2}{c}{OpenWebText} \\
\cmidrule(lr){3-4}\cmidrule(lr){5-6}
 & & Pythia-1B & OPT-1.3B & Pythia-1B & OPT-1.3B \\
\midrule
\multirow{3}{*}{FMLM}
 & 1 & 94.75 & 97.29 & 170.10 & 152.67 \\
 & 2 & 87.55 & 88.58 & 134.89 & 122.45 \\
 & 4 & 79.86 & 80.37 & 109.44 & 100.41 \\
\cmidrule(lr){1-6}
\multirow{3}{*}{Duo+DCD}
 & 1 & 829.59 & 849.31 & 4654.30 & 4401.95 \\
 & 2 & 435.40 & 453.47 & 905.38 & 792.78 \\
 & 4 & 254.97 & 260.45 & 246.35 & 219.22 \\
\cmidrule(lr){1-6}
\multirow{3}{*}{Duo+Di4C}
 & 1 & -- & -- & 1029.73 & 1061.16 \\
 & 2 & -- & -- & 736.75 & 728.92 \\
 & 4 & -- & -- & 252.60 & 251.94 \\
\cmidrule(lr){1-6}
\multirow{3}{*}{MDLM+SDTT}
 & 1 & 919.99 & 935.31 & 1024.45 & 1086.82 \\
 & 2 & 397.09 & 414.49 & 802.18 & 809.25 \\
 & 4 & 161.86 & 162.99 & 306.35 & 307.63 \\
\cmidrule(lr){1-6}
\multirow{3}{*}{MDLM+Di4C}
 & 1 & 1088.63 & 1120.40 & 1025.45 & 1067.07 \\
 & 2 & 601.01 & 626.70 & 702.27 & 695.89 \\
 & 4 & 266.63 & 273.17 & 216.24 & 215.50 \\
\midrule
\rowcolor{gray!12}
\textbf{Seq-Drifting$^{S}$} & \textbf{1} & \textbf{97.04} & \textbf{98.70} & \textbf{130.48} & \textbf{151.38} \\
\rowcolor{gray!12}
\textbf{Seq-Drifting$^{P}$} & \textbf{1} & \textbf{36.07} & \textbf{43.07} & \textbf{110.36} & \textbf{119.68} \\
\bottomrule
\end{tabular}
\end{table}

\subsection{Translation and summarization setup}
\label{sec:app_downstream_training}
For WMT14, the input contains a 64-token source sequence and the model generates a 64-token target
sequence. For XSum, the input contains up to 1024 source tokens and the model generates a 64-token
summary. We report corpus BLEU~\citep{papineni2002bleu} for WMT14 and ROUGE-1, ROUGE-2, and ROUGE-L~\citep{lin2004rouge} for XSum. 

For Seq-Drifting, we first train an autoregressive model from scratch on each task's
training split. In the small support-model setting, this model is a 12-layer pre-LN Transformer with hidden width $768$,
12 attention heads, FFN width $3072$, and the same tokenizer and context lengths used by the
corresponding experiment. It uses a causal attention mask and is trained with the standard
next-token cross-entropy objective. We then train Seq-Drifting using the frozen autoregressive model's
token distributions as its on-policy support signal, without initializing from a pretrained language
model. At evaluation time, we compare generated outputs against both the dataset ground-truth
references and samples generated by the autoregressive model, using the same task metrics in both
cases.

\subsection{Reasoning implementation details}
\label{sec:app_reasoning}
For mathematical reasoning, we train on GSM8K-Aug~\citep{deng2023implicit} and evaluate out-of-domain generalization on SVAMP and
ASDiv. Ground-truth solution sequences define the support set for the attraction objective. For logical reasoning, we use the binary (true/false) subset of ProofWriter, training on the combined D0–D5 training split and evaluating on the corresponding test split; unknown-labeled examples are excluded. Seq-Drifting uses a $0.5$B-parameter backbone with
the Qwen2.5 architecture. We compare it with autoregressive (AR), non-autoregressive (NAR), and latent-reasoning baseline Coconut~\citep{hao2025training}. We report answer accuracy and wall-clock time per sample for each benchmark in Table~\ref{tab:reasoning}.
\begin{itemize}
\item \textbf{Coconut} performs latent reasoning by feeding continuous thought
representations back into the model's hidden states instead of emitting explicit chain-of-thought tokens.
\end{itemize}
% TODO: Specify model initialization, support-set construction, training hyperparameters,
% and the training configurations of the baselines. Describe the GPQA training setup,
% evaluation split, prompting, and answer extraction. Document the timing hardware,
% batch size, generation limits, and averaging protocol.
% Add the dataset, architecture, and baseline citations when their bibliography keys are available.

%\subsection{Perplexity scaling}
%\label{sec:app_scaling_ppl}
%As the one-step generator's backbone grows, its unconditional generative perplexity keeps improving,
%so the drift objective benefits from scale rather than saturating at small sizes.
%Table~\ref{tab:scaling_ppl} reports gen-PPL across backbone sizes at length $128$, all in a single
%forward pass. \TODO{fill params and numbers.}
%\input{table_appendix_scaling_ppl}

\subsection{Training dynamics}
\label{sec:app_training}
\begin{figure}[t]
\centering
\includegraphics[width=0.48\linewidth]{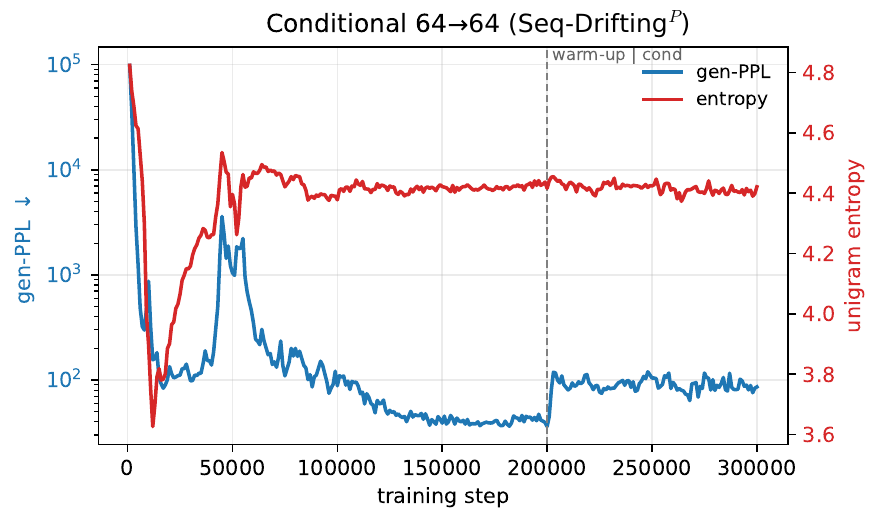}\hfill
\includegraphics[width=0.48\linewidth]{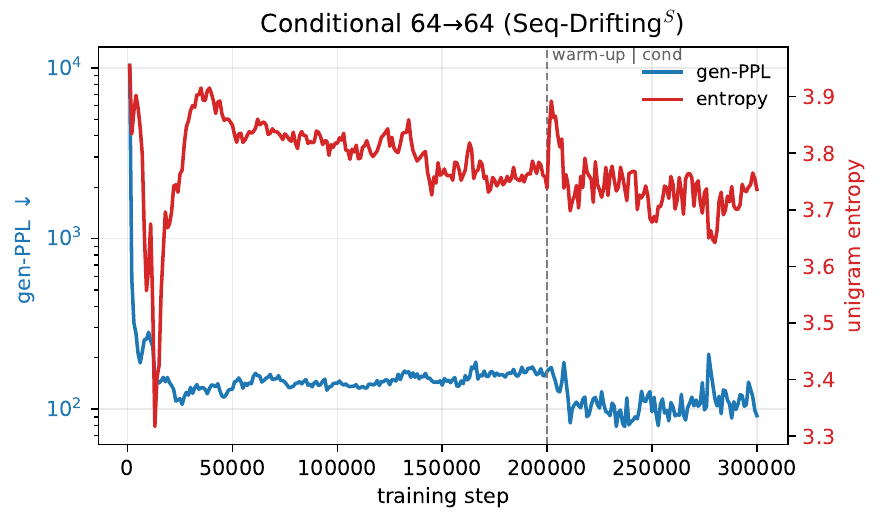}\\[2pt]
\includegraphics[width=0.48\linewidth]{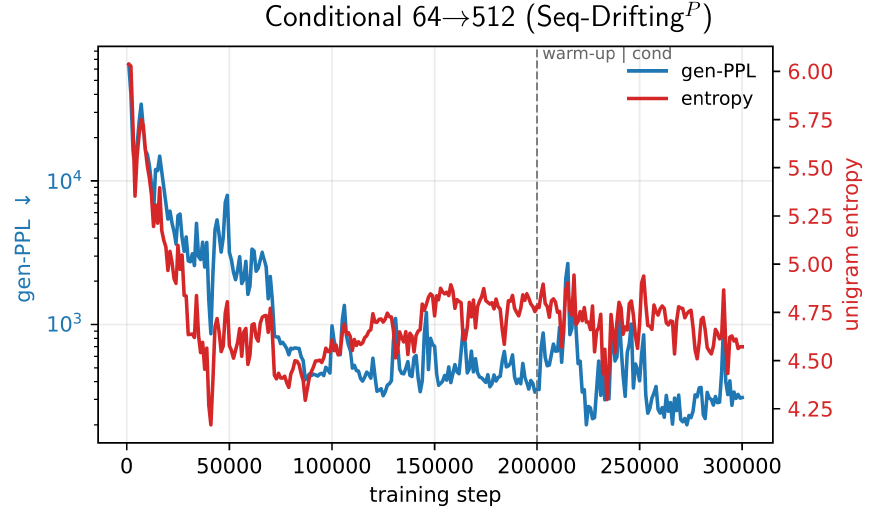}\hfill
\includegraphics[width=0.48\linewidth]{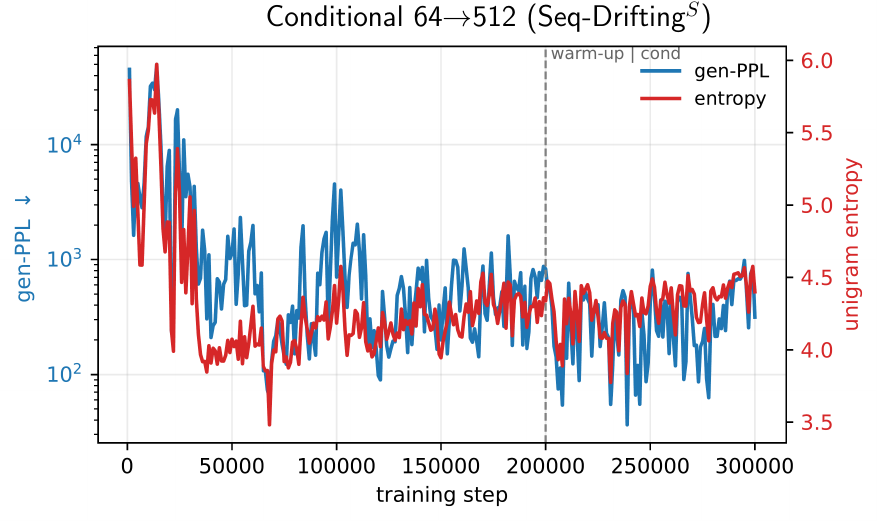}
\caption{\textbf{Training dynamics of the conditional runs.} Gen-PPL (log scale, left axis) and unigram
entropy (right axis) versus training step, for Seq-Drifting$^{P}$ $64{\to}64$ (top left), Seq-Drifting$^{S}$ $64{\to}64$
(top right), Seq-Drifting$^{P}$ $64{\to}512$ (bottom left), and Seq-Drifting$^{S}$ $64{\to}512$ (bottom right). All panels
show training through $300$k steps for both output lengths.
The dashed line marks the warm-up$\to$conditional switch at $200$k steps.}
\label{fig:training}
\end{figure}

Figure~\ref{fig:training} tracks Gen-PPL and unigram entropy over training for the conditional runs.
The first $200$k steps are the unconditional warm-up phase, during which the context is filled with a
constant token under normal attention. The drift reaches coherent English and drives Gen-PPL down by
orders of magnitude; real prefix conditioning then turns on and perplexity keeps falling.

\subsection{Off-policy positives}
\label{sec:app_offpolicy}
The \emph{off-policy} ablation of \S\ref{sec:exp_ablation} constructs support sets from corpus retrieval instead of language-model predictions. We
precompute a memory bank of the token $n$-gram embeddings of a set of real sequences, and at each
generated position we take as the support set $\mathcal{S}_t$ the tokens that real sequences place at
the position where they best match the current generation, ranked by the position-aligned min-$n$-gram
distance of \S\ref{sec:tdd_support}. Everything downstream, the angular attraction, the two repulsions,
and the geodesic update, is unchanged. The one difference from the on-policy signal is that these
targets are \emph{fixed} by the data and do not follow the generator.

\begin{figure}[t]
\centering
\includegraphics[width=0.56\linewidth]{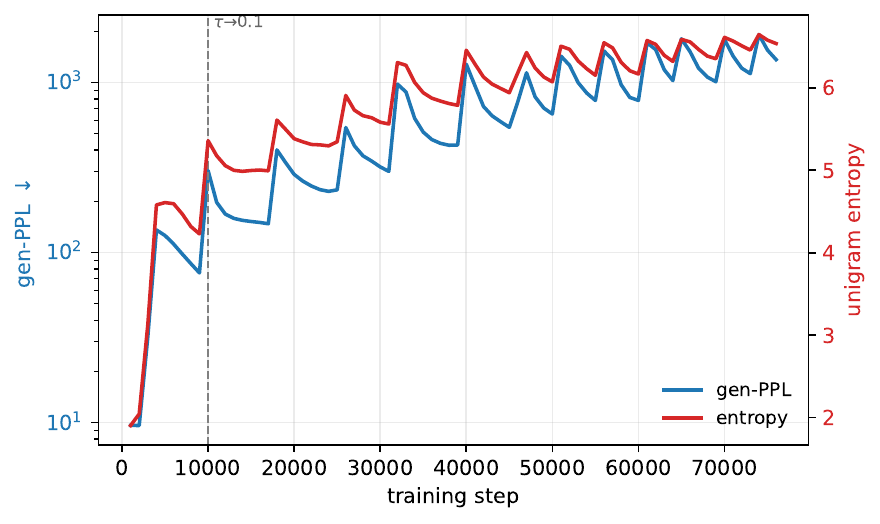}
\caption{\textbf{Off-policy training dynamics} (unconditional, length $128$). Gen-PPL (log scale, left
axis) and unigram entropy (right axis) versus training step for the off-policy positives ablation. The
dashed line marks the end of the temperature warm-up at $10$k steps, where $\tau$ reaches $0.1$ and the
run begins to oscillate and degrade.}
\label{fig:offpolicy}
\end{figure}

\paragraph{Observed training behavior.} Figure~\ref{fig:offpolicy} tracks this run at length $128$.
During the temperature warm-up, over which the attraction temperature $\tau$ anneals from $1.0$ to
$0.1$ across the first $10$k steps, Gen-PPL falls as the soft, mode-covering attraction pulls the batch
toward real $n$-grams. Once $\tau$ reaches its sharp steady value at $10$k, each generation commits
hard to its single nearest real $n$-gram. Because those targets are fixed and do not adapt, the sharp
commitment lets the anti-collapse repulsion over-scatter the batch faster than the fixed attraction can
pull it back, which sets up a limit cycle. The repulsion over-scatters the batch, so entropy and
Gen-PPL spike. The samples are then far apart, the repulsion weakens, and the attraction pulls them back
toward real tokens, so entropy and Gen-PPL fall. Once re-consolidated the repulsion strengthens again
and the cycle repeats, which is the sawtooth in Figure~\ref{fig:offpolicy}. The cycle does not close,
each swing ends higher than the last, so Gen-PPL and entropy trend upward overall and the run moves toward
higher-entropy text with reduced global coherence.
The on-policy support model avoids this
this because it scores the generation's own prefix, so its targets follow the generator and the
attraction stays balanced against the repulsion.

\subsection{Qualitative samples}
\label{sec:app_case}
The following examples are produced in a single forward pass and truncated for display.

\samplebox{Seq-Drifting$^{P}$}{
entropy: 4.30 Gen-PPL: 36.01
}{
It will provide an excellent overview of a wide variety of topics: "Innovations \& Challenges of the Future," "A History of Technology" and, the latest developments
and advancements that the technology has in general and the development and future capabilities in
specific applications. This is one of our largest conferences ever organized for industry-leading
attendees in North American technology industry, which will give you more insight as you go to work
with technology.

About the Author:

Klaus K. Krieg was awarded his doctor, \ldots
}
\vspace{6pt}
\samplebox{Seq-Drifting$^{S}$}{
entropy: 4.21 Gen-PPL: 93.81
}{
A spokeswoman at the Brown's Downing Street offices said: "There it would not happen here at home,
will be able to make up his mind whether that's right another very me. i guess he has been very,
but if he did have been opportunity, if he has been very very me. the so far as well ... and this
time around there, if he did. i think terms people who knows about what happened on afternoon,
\ldots
}
%\vspace{6pt}
%\samplebox{Seq-Drifting$^{S}$ (length=1024)}{
%}{
%This post originally published onHAM Huffington Post strives are AmberTRUMP-itbart pristine: after OprahGrid the Friday president on the Wednesday morning to recalled on the afternoon, he've been a fraternity and first has beenOAD: The same: Theitage: a Ohio: The Ohio: a voted: The Republican- not fullest on a nominating and he ultr and the Tea and have been on on the same, thetorn. The guy- he perk to an jury and he legislator, a Presidency by a lbs of aemption for he. to been on theshe sprawling, aninguished- a antidepress of theourgeois
%}
\vspace{6pt}
\samplebox{Seq-Drifting WMT}{}{
\textbf{Source:} Wir sind stolz auf unsere Leistung, aber wir wollen jedes Spiel gewinnen.

\textbf{Reference:} We are proud of our performance, but we want to win every game.

\textbf{Prediction:} We are proud of our performance, but we want to win every game.
}
\vspace{6pt}
\samplebox{Seq-Drifting XSum}{}{
\textbf{Source:} Police said officers from the North West counter terrorism unit searched an
address on Peakdale Avenue, Crumpsall, Manchester on Friday. The suspect, 26, was arrested the
following day on suspicion of offences under the Terrorism Act. Police would not comment on the
nature of the alleged offence but said it believed it caused "no threat" to the community.

\textbf{Reference:} A man is being held in Manchester on suspicion of terrorism offences.

\textbf{Prediction:} A man has been arrested on suspicion of terrorism offences
}
\vspace{6pt}
\samplebox{Seq-Drifting Math Reasoning}{}{
\textbf{Question:} A group of students are preparing for a water balloon fight. Each student needs 5
balloons. If there are 12 students, how many balloons are needed in total?

\textbf{Prediction:} \texttt{<<5*12=60>>}\\
\texttt{\#\#\#\# 60}
}

\vspace{6pt}
\samplebox{Seq-Drifting ProofWriter}{}{
\textbf{Question:} Theory: Bob is kind. Bob is nice. Bob is white. Charlie is kind. Charlie is white.
Erin is red. Erin is rough. Harry is blue. Harry is kind. Harry is red. Blue people are nice. All white
people are red. If someone is white and blue then they are nice. All rough people are red. If someone is
smart then they are blue. If someone is kind then they are red. If someone is nice then they are rough.
If someone is red then they are smart. If someone is red and rough then they are nice.

\textbf{Query:} Bob is not nice. Give the proof in order and finish with \texttt{\#\#\#\# true} or
\texttt{\#\#\#\# false}.

\textbf{Prediction:} Fact: Bob is nice.\\
\texttt{\#\#\#\# false}
}

\end{document}